\documentclass[12pt]{article}

\usepackage{graphicx} 
\usepackage{multicol}
\usepackage{setspace}

\usepackage[top=1in, bottom=1in, left=1in, right=1in]{geometry}

\usepackage{algorithm}
\usepackage{algorithmic}

\usepackage{hyperref}

\usepackage{latexsym}
\usepackage{amssymb}

\usepackage{graphicx}
\usepackage{rotating}
\usepackage{url}
\usepackage{epsfig}

\usepackage{amsmath}
\usepackage{amssymb}
\usepackage{bm}

\usepackage[table]{xcolor}

\newcommand{\g}{\, | \,}

\newcommand{\expectq}[1]{\mathbb{E}_q\left[#1\right]}

\newcommand{\expect}[1]{\mathbb{E}\left[#1\right]}

\newcommand{\partl}[2]{\frac{\partial #1}{\partial #2}}
\newcommand{\partll}[2]{\frac{\partial^2 #1}{\partial {#2}^2}}
\newcommand{\partlll}[3]{\frac{\partial^2 #1}{\partial {#2}{\partial {#3}^T}}}

\newcommand{\myeq}[1]{Equation~\ref{equation:#1}}
\newcommand{\myappendix}[1]{Appendix~\ref{appendix:#1}}
\newcommand{\mytable}[1]{Table~\ref{table:#1}}
\newcommand{\mysec}[1]{Section~\ref{section:#1}}
\newcommand{\myfig}[1]{Figure~\ref{figure:#1}}

\usepackage{natbib}

\title{The Issue-Adjusted Ideal Point Model}

\author{Sean Gerrish \\
  Princeton University \\
  35 Olden Street\\
  Princeton, NJ 08540\\
  {\tt sgerrish@cs.princeton.edu}
\and
David M. Blei\\
Princeton University\\
35 Olden Street\\
Princeton, NJ 08540\\
{\tt blei@cs.princeton.edu}}

\date{}

\usepackage{titlesec}

\bibpunct{(}{)}{;}{a}{}{,}
\begin{document}
\maketitle

\newpage

David M. Blei is a Computer Science Professor, Princeton Computer
Science Department, Princeton, NJ, 08540; and Sean M. Gerrish is a
computer science graduate student, Princeton University, Princeton, NJ
08540. This work was partially supported by ONR N00014-11-1-0651, NSF
CAREER 0745520, AFOSR FA9550-09-1-0668, the Alfred P. Sloan
Foundation, and a grant from Google.

\newpage

\begin{abstract}
  We develop a model of issue-specific voting behavior.  This model
  can be used to explore lawmakers' personal voting patterns of voting
  by issue area, providing an exploratory window into how the language
  of the law is correlated with political support.  We derive
  approximate posterior inference algorithms based on variational
  methods.  Across 12 years of legislative data, we demonstrate both
  improvement in heldout prediction performance and the model's
  utility in interpreting an inherently multi-dimensional space.
\end{abstract}
Key words: Item response theory, Probabilistic topic model, Variational inference, Legislative voting

\titleformat{\section}
  {\normalfont\large\uppercase}{\large \thesection.}{.5em}{}

\titleformat{\subsection}
  {\normalfont}{\thesubsection.}{.5em}{}

\section{Introduction}

Legislative behavior centers around the votes made by lawmakers.
These votes are captured in \textit{roll call data}, a matrix with
lawmakers in the rows and proposed legislation in the columns.  We
illustrate a sample of roll call votes for the United States Senate in
\myfig{roll_call_table}.
\begin{figure}[b]
  \center
  \textbf{Example roll call votes}
  \begin{tabular}{cccccc}
   \hline
   \hline
   \small Lawmaker & \multicolumn{5}{c}{\small Item of legislation} \\
   \hline
   \small Bill & \small  \small S. 3930 & \small  H.R. 5631 & \small  H.R. 6061 & \small  H.R. 5682 & \small  S. 3711 \\
   \small Mitch McConnell (R) & \small  \small Yea & \small  Yea & \small  Yea & \small  Yea & \small  Yea \\
   \small Olympia Snowe (R) & \small   & \small  Yea & \small  Yea & \small  Yea & \small  Nay \\
   \small John McCain (R) & \small  Yea & \small  Yea & \small  Yea & \small  Yea & \small  Yea \\
   \small Patrick Leahy (D) & \small  Nay & \small  Yea & \small  Nay & \small  Nay & \small  Nay \\
   \small Paul Sarbanes (D) & \small  Nay & \small  Yea & \small  Nay & \small  Yea & \small  Nay \\
   \small Debbie Stabenow (D) & \small  Yea & \small  Yea & \small  Yea & \small  Yea & \small  Yea \\
   \hline
 \end{tabular}
 \caption{A sample roll-call matrix illustrating lawmakers' votes on
   items of legislation.  These votes are from the Senate in the 109th
   Congress (2005-2006).  The party of each Senator -- (D)emocrat or
   (R)epublican -- is provided in parentheses. This matrix is
   sometimes incomplete (see Snowe's vote on S. 3930, for example). }
  \label{figure:roll_call_table}
\end{figure}

\begin{figure}
  \begin{center}
    \includegraphics[width=\textwidth]{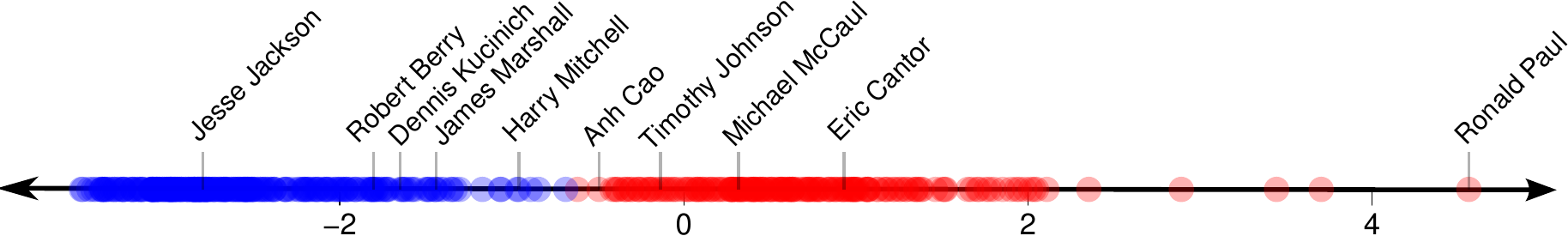}
 \end{center}
 \caption{Traditional ideal points separate Republicans (red) from
   Democrats (blue).}
  \label{figure:classic_ideal_points}
\end{figure}

The seminal work of \cite{poole:1985} introduced the \textit{ideal
  point model}, using roll call data to infer the latent political
positions of the lawmakers.  The ideal point model is a latent factor
model of binary data and an application of item-response
theory~\citep{lord:1980} to roll call data. It gives each lawmaker a
latent political position along a single dimension and then uses these
points (called the ideal points) in a model of the votes.  (Two
lawmakers with the same position will have the same probability of
voting in favor of each bill.)  From roll call data, the ideal point
model recovers the familiar division of Democrats and Republicans. See
\myfig{classic_ideal_points} for an example.

\begin{figure}
  \begin{tabular}{|c|}
    \hline
    { } \\
    \includegraphics[width=\textwidth]{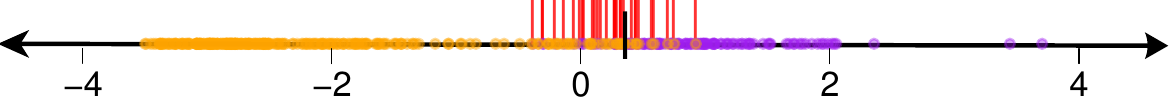} \\
    Incorrect votes by classic ideal point \\
  \hline
    { } \\
  \includegraphics[width=\textwidth]{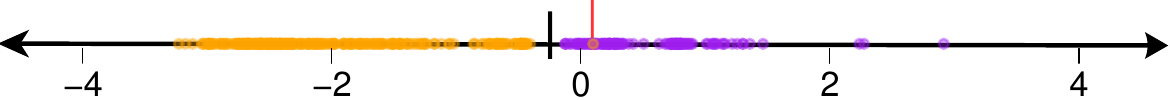} \\
  Incorrect votes by issue-adjusted ideal point \\
  \hline
  \end{tabular}
  \caption{Classic ideal points (top) represent votes incorrectly when
    lawmakers hold issue-specific opinions, while issue-adjusted ideal
    points (bottom) can account for this.  Classic ideal points assume
    that lawmakers hold fixed positions, while issue-adjusted ideal
    points allow their positions to change by issue. Each point above
    is the ideal point of a lawmaker voting on an act
    \emph{Recognizing the significant accomplishments of AmeriCorps}
    [and raising community service] (H.R. 1338 in Congress 111);
    orange points represent lawmakers who voted ``Yea'', and violet
    points represent lawmakers who voted ``Nay'' on this bill.  The
    theory behind classic ideal points assumes that lawmakers' votes
    on a bill can be described by their side of the cut point (black
    vertical line). Red lines mark lawmakers whose votes were
    incorrectly predicted with each model.}
  \label{figure:bad_ideal_points}
\end{figure}

Ideal point models can capture the broad political structure of a body
of lawmakers, but they cannot tell the whole story.  We illustrate
this with votes on a bill in \myfig{bad_ideal_points}.  This figure
shows lawmaker's ideal points for their votes on an act
\emph{Recognizing the significant accomplishments of AmeriCorps},
H.R. 1338 in Congress 111.  In this figure, ``Yea'' votes are colored
orange, while ``Nay'' votes are violet; a classic ideal point model
predicted that votes to the \emph{right} of the vertical line were
``Nay'' while those to the left were ``Yea''.  Out of four hundred
eight votes on this bill modeled by an ideal point model, thirty-one
of these were modeled incorrectly.

Sometimes these votes are incorrectly predicted because of stochastic
circumstances surrounding lawmakers and bills.  More often, however,
these votes can be explained because lawmakers are not
one-dimensional: they each hold positions on different issues. For
example, Ronald Paul, a Republican representative from Texas, and
Dennis Kucinich, a Democratic representative from Ohio, hold
consistent political opinions that an ideal point model systemically
gets incorrect.  Looking more closely at these errors, we would see
that Paul differs from a typical Republican when it comes to foreign
relations and social issues; Kucinich differs from a usual Democrat
when it comes to foreign policy.

The problem is that classical ideal point models place each lawmaker
in a single political position, but a lawmaker's vote on a bill has to
do with a number of factors---her political affiliation, the content
of the proposed legislation, and her political position \textit{on
  that content}.  While classical ideal point models can capture the
main regularities in lawmakers' voting behavior, they cannot predict
when and how a lawmaker will vote differently than we expect.

In this paper, we develop the \textit{issue-adjusted ideal point
  model}, a model that captures issue-specific deviation in lawmaker
behavior.  We place the lawmakers on a political spectrum and identify
how they deviate from their position as a function of specific issues.
This results in inferences like those illustrated in Figure 3.  An
important component of our model is that we use the text of the
proposed bills to encode which issues they are about.  (We do this
through a probabilistic topic model~\citep{blei:2003}.)  Unlike other
attempts at developing multi-dimensional ideal point
models~\citep{jackman:2001}, our approach explicitly ties the
additional dimensions to the political discussion at hand.

By incorporating issues, we can model the AmeriCorp bill above much
better than we could with classic ideal points (see
\myfig{bad_ideal_points}). By recognizing that this bill is about
\emph{social services}, and by modeling lawmakers' positions on this
issue, we are able to predict all but one of the lawmakers' votes
correctly.  This is because we can learn to differentiate between
lawmakers who are conservative and lawmakers who are conservative on
\emph{social services}.  For example, the issue-adjusted model tells
us that, while Doc Hastings (Republican of Washington) is considered
more conservative than Timothy Johnson (Republican of Illinois) in the
ideal point model, Hastings is much more liberal on social issues than
Johnson---hence, he will more often generally side with Democrats on
those votes.

In the following sections, we describe our model and develop efficient
approximate posterior inference algorithms for computing with it.  To
handle the scale of the data we want to study, we replace the usual
MCMC approach with a faster variational inference algorithm.  We then
study 12 years of legislative votes from the U.S. House of
Representatives and Senate, a collection of 1,203,009 votes.  We show
that our model gives a better fit to the data than a classical ideal
point model and demonstrate that it provides an interesting
exploratory tool for analyzing legislative behavior.

\paragraph{Related work.}

%

Item response theory (IRT) has been used for decades in political
science \citep{clinton:2004,martin:2002,poole:1985}; see
\cite{fox:2010} for an overview, \cite{enelow:1984} for a historical
perspective, and \cite{albert:1992} for Bayesian treatments of the
model.  Some political scientists have used higher-dimensional ideal
points, where each legislator is described by a vector of ideal points
$\bm x_u \in \mathbb{R}^K$ and each bill polarization $\bm a_d$ (i.e.,
how divisive it is) takes the same dimension $K$
\cite{heckman:1996}. The probability of a lawmaker voting ``Yes'' is
$\sigma(\bm x_u^T \bm a_d + b_d)$ (we describe these assumptions
further in the next section).  The principle component of ideal points
explains most of the variance and explains party affiliation.
However, other dimensions are not attached to issues, and interpreting
beyond the principal component is painstaking \citep{jackman:2001}.

At the minimum, this painstaking analysis often requires careful study
of the original roll-call votes or study of lawmakers' ideal-point
neighbors.  The former obviates an IRT model, since we cannot make
inferences from model parameters alone; while the latter begs the
question, since it assumes we know in the first place how lawmakers
vote on different issues. The model we discuss in this paper is
intended to address this problem by providing interpretable
multi-dimensional ideal points.  Through posterior inference, we can
estimate each lawmaker's political position and how it changes on a
variety of concrete issues.

The model we will outline takes advantage of recent advances in
content analysis, which have received increasing attention because of
their ability to incorporate large collections of text at a relatively
small cost (see \cite{grimmer:2012} for an overview of these
methods).  For example, \cite{quinn:2006b} used text-based methods to
understand how legislators' attention was being focused on different
issues, to provide empirical evidence toward answering a variety of
questions in the political science community.

We will draw heavily on content analytic methods in the machine
learning community, which has developed useful tools for modeling both
text and the behavior of individuals toward items. Recent work in this
community has provided joint models of legislative text and
votes. \cite{gerrish:2011} aimed to predict votes on bills which had
not yet received any votes.  This model fitted predictors of each
bill's parameters using the bill's text, but the underlying voting
model was still one-dimensional--it could not model individual votes
better than a one-dimensional ideal point model.  In other work,
\cite{wang:2010} developed a Bayesian nonparametric model of votes and
text over time.  Both of these models have different purposes from the
model presented here; neither addresses individuals' affinity toward
different types of bills.

The issue-adjusted model is conceptually more similar to recent models for
content recommendation.  Specifically, \cite{wang:2011} describe a
method to recommend academic articles to users of a service based on
what they have already read, and \cite{agarwal:2010} proposed a
similar model to match users to other items (i.e., Web content).  Our
model is related to these approaches, but it is specifically designed
to analyze political data.  These works, like ours, model users'
affinities to items.  However, neither of them employ the notion of
the orientation of an item (i.e., the political orientation of a bill)
or that the users (i.e., lawmakers) have a position on a this
spectrum.  These are considerations which are required when analyzing
political roll call data.



\section{The Issue-Adjusted Ideal Point Model}
\label{section:exceptional_model}

We first review ideal point models of legislative roll call data and
discuss their limitations.  We then present our model, the
\textit{issue-adjusted ideal point model}, that accounts for how
legislators vote on specific issues.

\subsection{Modeling Political Decisions with Ideal Point Models}

Ideal point models are latent variable models that have become a
mainstay in quantitative political science.  These models are based on
item response theory, a statistical theory that models how members of
a population judge a set of items (see \cite{fox:2010} for an
overview).  Applied to voting records, ideal point models place
lawmakers on an interpretable political spectrum.  They are widely
used to help characterize and understand historical legislative and
judicial decisions~\citep{clinton:2004,poole:1985,martin:2002}.

One-dimensional ideal point models posit an \textit{ideal point} $x_u
\in \mathbb{R}$ for each lawmaker $u$.  Each bill $d$ is characterized
by its \textit{polarity} $a_d$ and its \textit{popularity} $b_d$. (The
polarity is often called the ``discrimination'', and the popularity is
often called the ``difficulty''; polarity and popularity are more
accurate terms.)  The probability that lawmaker $u$ votes ``Yes'' on
bill $d$ is given by the logistic regression
\begin{equation}
  \label{equation:trad_ipm}
  p(v_{ud} = \textrm{yes} \g x_u, a_d, b_d) =
  \sigma(x_u a_d + b_d),
\end{equation}
where $\sigma(s) = \exp(s) / (1 + \exp(s))$ is the logistic
function. (A probit function is sometimes used instead of the
logistic.  This choice is based on an assumption in the underlying
model, but it has little empirical effect in legislative ideal point
models.)
When the popularity of a bill $b_d$ is high, nearly everyone votes
``Yes''; when the popularity is low, nearly everyone votes ``No''.
When the popularity is near zero, the probability that a lawmaker
votes ``Yes'' is determined primarily by how her ideal point $x_u$
interacts with bill polarity $a_d$.





In Bayesian ideal point modeling, the variables $a_d$, $b_d$, and
$x_u$ are usually assigned standard normal
priors~\citep{clinton:2004}.  Given a matrix of votes $\bm{v} =
\{v_{ud}\}$, we can estimate the posterior expectation of the ideal
point of each lawmaker $\expect{x_u \g \bm{v}}$.
Figure~\ref{figure:classic_ideal_points} illustrates ideal points
estimated from votes in the U.S. House of Representatives from
2009-2010.  The model has clearly separated lawmakers by their
political party (color) and provides an intuitive measure of their
political leanings.


\subsection{Limitations of Ideal Point Models}

The ideal point model fit to the House of Representatives from
2009-2010 correctly models 98\% of all lawmakers' votes on training
data.  (We correctly model an observed vote if its probability under
the model is bigger than $1/2$.)  But it fits some lawmakers better
than others.  It only predicts 83.3\% of Baron Hill's (D-IN) votes and
80.0\% of Ronald Paul's (R-TX) votes.  Why is this?

To understand why, we look at how the ideal point model works.  The
ideal point model assumes that lawmakers are ordered, and that each
bill $d$ splits them at a \emph{cut point}.  The cut point is a
function of the bill's popularity and polarity, $-b_d/a_d$.  Lawmakers
with ideal points $x_u$ to one side of the cut point are more likely
to support the bill; lawmakers with ideal points to the other side are
more likely to reject it.  The issue with lawmakers like Paul and
Hill, however, is that this assumption is too strong---their voting
behavior does not fit neatly into a single ordering.  Rather, their
location among the other lawmakers changes with different bills.
\begin{figure}
 \begin{center}
   \includegraphics[width=0.85\textwidth]{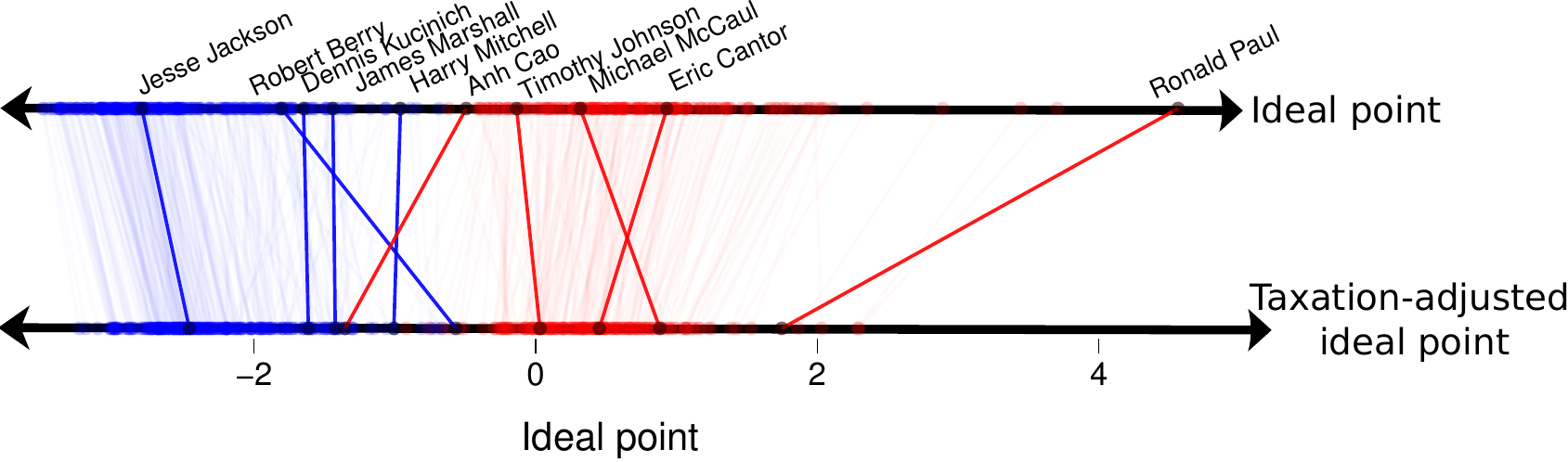}

  \vspace{0.25in}

  \includegraphics[width=0.85\textwidth]{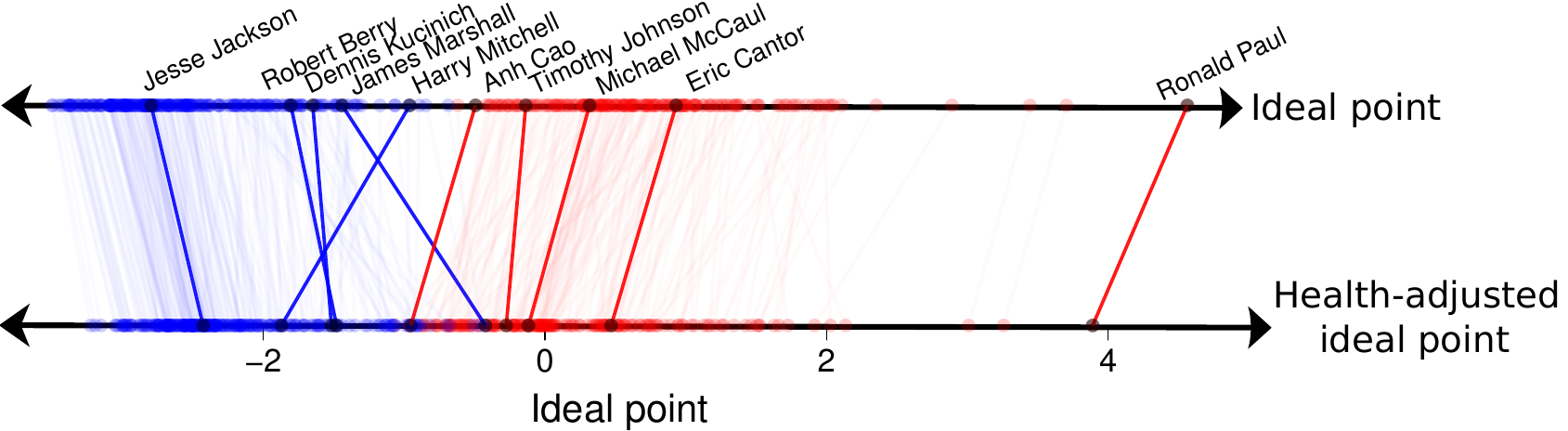}
\end{center}
\caption{In a traditional ideal point model, lawmakers' ideal points
  are static.  In the issue-adjusted ideal point model, lawmakers'
  ideal points change when they vote on certain issues, such as
  \emph{taxation} (top panel) and \emph{health} (bottom panel).  A
  line segment connects select lawmakers' ideal points (top row of
  each panel) to their issue-adjusted ideal points (bottom row of each
  panel).  Unlabeled lawmakers are illustrated by the remaining, faint
  line segments.  We have colored Democrats blue and Republicans red.}
  \label{fig:moving_ideal_points}
\end{figure}

However, there are still patterns to how they vote.  Paul and Hill
vote consistently within individual areas of policy, such as foreign
policy or education, though their voting on these issues diverges from
their usual position on the political spectrum.  In particular, Paul
consistently votes against United States involvement in foreign
military engagements, a position that contrasts with other
Republicans. Hill, a ``Blue Dog'' Democrat, is a strong supporter of
second-amendment rights, opposes same-sex adoption, and is wary of
government-run health care---positions that put him at odds with many
other Democrats.  Particularly, the ideal point model would predict
Paul and Hill as having muted positions along the classic left-right
spectrum, when in fact they have different opinions about
certain issues than their fellow legislators.



We refer to voting behavior like this as \emph{issue voting}.  An
issue is any federal policy area, such as ``financial
regulation,'' ``foreign policy,'' ``civil liberties,'' or
``education,'' on which lawmakers are expected to take positions.
Lawmakers' positions on these issues may diverge from their
traditional left/right stances, but traditional ideal point models
cannot capture this.  Our goal is to develop an ideal point model that
allows lawmakers to deviate, depending on the issue under discussion,
from their usual political position.

Figure~\ref{fig:moving_ideal_points} illustrates the kinds of
hypotheses our model can make.  Each panel represents an issue;
\textit{taxation} is on the top, and \textit{health} is on the bottom.
Within each panel, the top line illustrates the ideal points of
various lawmakers---these represent the relative political positions
of each lawmaker for most issues.  The bottom line illustrates the
position adjusted for the issue at hand.  For example, the model
posits that Charles Djou (Republican representative for Hawaii) is
more similar to Republicans on \emph{taxation} and more similar to
Democrats on \emph{health}, while Ronald Paul (Republican
representative for Texas) is more Republican-leaning on \emph{health}
and less extreme on \emph{taxation}.  Posterior estimates like this
give us a window into voting behavior that is not available to classic
ideal point models.

\subsection{Issue-adjusted Ideal Points}




The issue-adjusted ideal point model is a latent variable model of
roll call data.  As with the classical ideal point model, bills and
lawmakers are attached to popularity, polarity, and ideal points.  In
addition, the text of each bill encodes the issues it discusses and,
for each vote, the ideal points of the lawmakers are adjusted
according to those issues.  (We obtain issue codes from text by using
a probabilistic topic model.  This is described below in
Section~\ref{section:lda}.)

In more detail, each bill is associated with a popularity $a_d$ and
polarity $b_d$; each lawmaker is associated with an ideal point $x_u$.
Assume that there are $K$ issues in the political landscape, such as
\textit{finance}, \textit{taxation}, or \textit{health care}.  Each
bill contains its text $\bm{w}_d$, a collection of observed words,
from we which we derive a $K$-vector of \textit{issue proportions}
$\bm{\theta}(\bm{w}_d)$.  The issue proportions represent how much
each bill is about each issue.  A bill can be about multiple issues
(e.g., a bill might be about the tax structure surrounding health
care), but these values will sum to one.  Finally, each lawmaker is
associated with a real-valued $K$-vector of \textit{issue adjustments}
$\bm{z}_u$.  Each component of this vector describes how his or her
ideal point changes as a function of the issues being discussed.  For
example, a left-wing lawmaker may be more right wing on defense; a
right-wing lawmaker may be more left wing on social issues.

For the vote on bill $d$, we linearly combine the issue proportions
$\bm{\theta}(\bm{w}_d)$ with each lawmaker's issue adjustment
$\bm{z}_u$ to give an adjusted ideal point $x_u + \bm{z}_u^\top
\bm{\theta}(\bm{w}_d)$.  The votes are then modeled with a logistic
regression,
\begin{equation}
  \label{equation:ia-ipm}
  p(v_{ud} | a_d, b_d, \bm{z}_u, x_u, \bm{w}_d) =
  \sigma \left( (x_u + \bm{z}_u^\top \bm{\theta}(\bm{w}_d)
    ) a_d + b_d \right).
\end{equation}
We put standard normal priors on the ideal points, polarity, and
popularity variables.  We use Laplace priors for the issue adjustments
$\bm z_{u}$,
\begin{equation*}
  p(z_{uk} \g \lambda_1) \propto \exp\left( - \lambda_1 || z_{uk} ||_1
\right).
\end{equation*}
Using MAP inference, this finds sparse adjustments.  With full
Bayesian inference, it finds nearly-sparse adjustments.  Sparsity is
desirable for the issue adjustments because we do not expect each
lawmaker to adjust her ideal point $x_u$ for every issue; rather, the
issue adjustments are meant to capture the handful of issues on which
she does diverge.

\begin{figure}
  \begin{center}
    \includegraphics[width=0.75\textwidth]{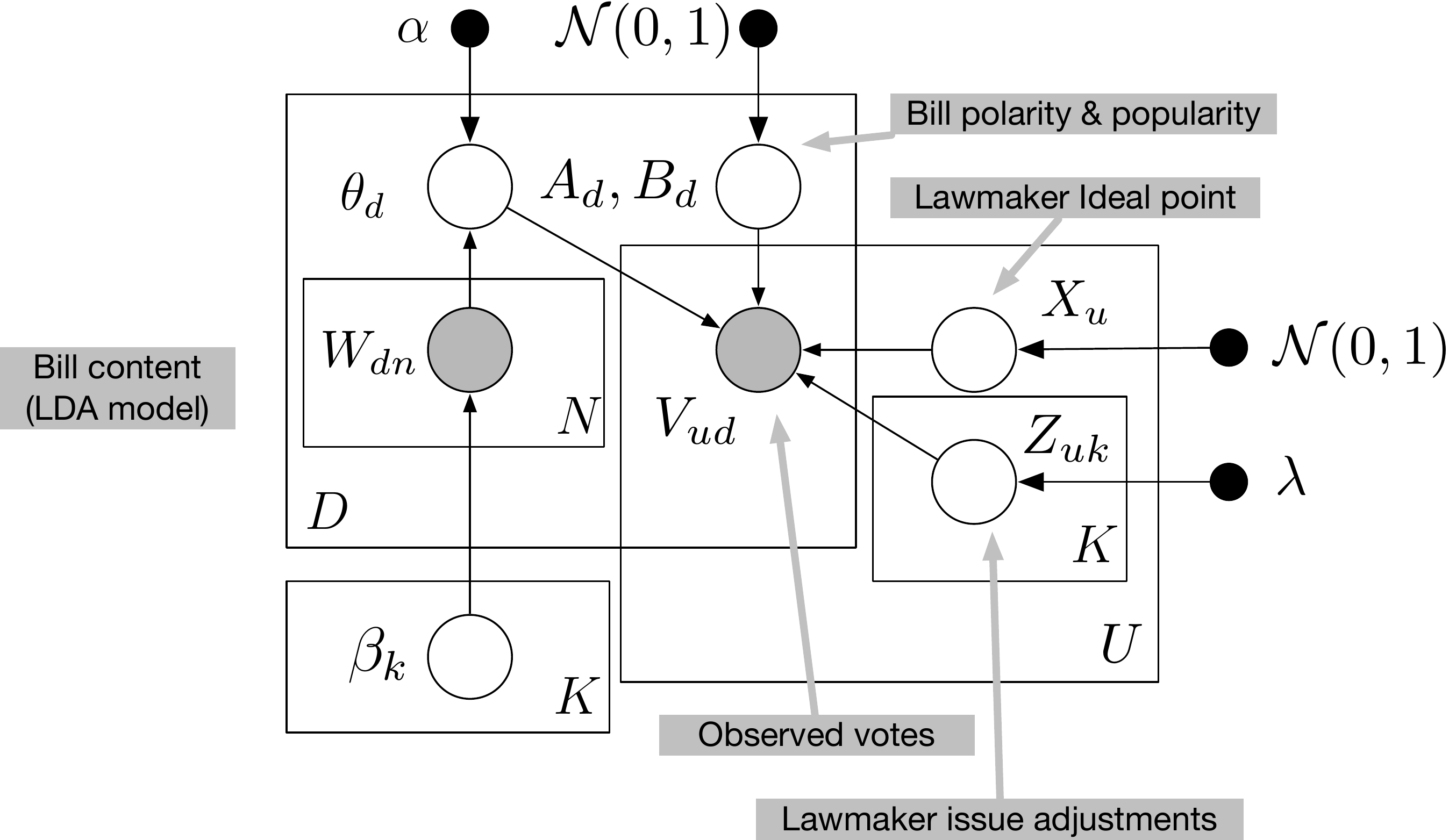}
  \end{center}

  \caption{\label{figure:legis_gm} A graphical model for the
    issue-adjusted ideal point model, which models votes $v_{ud}$ from
    lawmakers and legislative items.  Lawmakers' positions are
    determined by $x_u$ and $z_u$, a $k$-vector which interacts with
    bill-specific issue mixtures $\theta_d$ (also $k$-vectors).  Issue
    mixtures are fit from text using labeled latent Dirichlet
    allocation.  As with ideal points models, $a_d$ and $b_d$ are
    bill-specific variables describing the bill's polarization and
    popularity.}
\end{figure}

Suppose there are $U$ lawmakers, $D$ bills, and $K$ issues.  The
generative probabilistic process for the issue-adjusted ideal point
model is the following.
\begin{enumerate}
\item For each user $u \in \{1, \ldots, U\}$:
  \begin{enumerate}
  \item Draw ideal points $x_u \sim {\cal N}(0,1)$.
  \item Draw issue adjustments $z_{uk} \sim
    \textrm{Laplace}(\lambda_1)$ for each issue $k \in \{1, \ldots,
    K\}$.
  \end{enumerate}
\item For each bill $d \in \{1, \ldots, D\}$:
  \begin{enumerate}
  \item Draw polarity $a_d \sim {\cal N}(0,1)$.
  \item Draw popularity $b_d \sim {\cal N}(0,1)$.
  \end{enumerate}
\item Draw vote $v_{ud}$ from \myeq{ia-ipm} for each user/bill pair,
  $u \in \{1, \ldots, U\}$ and $d \in \{1, \ldots, D\}$.
\end{enumerate}
Figure~\ref{figure:legis_gm} illustrates the graphical model.  Given
roll call data and bill texts, we can use posterior expectations to
estimate the latent variables.  For each lawmaker, these are the
expected ideal points and per-issue adjustments; these are the
posterior estimates we illustrated in
Figure~\ref{fig:moving_ideal_points}.  For each bill, these are the
expected polarity and popularity.

We consider a simple example to better understand this model.  Suppose
a bill $d$ is only about \emph{finance}.  This means that $\bm
\theta(\bm w_d)$ has a one in the \emph{finance} dimension and zero
everywhere else.  With a classic ideal point model, a lawmaker $u$'s
ideal point $x_u$ gives his position on every bill, regardless of the
issue.  With the issue-adjusted ideal point model, his \emph{effective
  ideal point} for this bill is $x_u + z_{u,\mbox{\tiny Finance}}$,
adjusting his position based on the bill's content.  The adjustment
$z_{u,\mbox{\tiny Finance}}$ might move him to the right or the left,
capturing an issue-dependent change in his ideal point.

In the next section we will describe a posterior inference algorithm
that will allow us to estimate $x_u$ and $\bm z_{u}$ from lawmakers'
votes.  An eager reader can scan ahead to browse these effective ideal
points for Ron Paul, Dennis Kucinich, and a handful of other lawmakers
in Figure~\ref{figure:issue_improvements_ideals}.  This figure shows
the posterior mean of issue-adjusted ideal points that have been
inferred from votes about \emph{finance} (top) and votes about
\emph{congressional sessions} (bottom).

In general, a bill might involve several issues; in that case the
issue vector $\bm \theta(\bm w_d)$ will include multiple positive
components.  We have not yet described this important function, $\bm
\theta(\bm w_d)$, which codes a bill with its issues.  We describe
that function in \mysec{lda}.  First we discuss the relationship
between the issue adjusted model and other models of political science
data.

\subsection{Relationship to Other Models of Roll-call Data}
\label{section:relationship}

The issue-adjusted ideal point model recovers the classical ideal
point model if all of the adjustments (for all of the lawmakers) are
equal to zero.  In that case, as for the classical model, each bill
cuts the lawmakers at $-b_d/a_d$ to determine the probabilities of
voting ``yes.''  With non-zero adjustments, however, the model asserts
that the relative positions of lawmakers can change depending on the
issue.  Different bill texts, through the coding function
$\bm{\theta}(\bm{w}_d)$, will lead to different orderings of the
lawmakers.  Again, Figure~\ref{fig:moving_ideal_points} illustrates
these re-orderings for idealized bills, i.e., those that are only
about taxation or healthcare.

Issue adjusted models are an interpretable multidimensional ideal
point model.  In previous variants of multidimensional ideal point
models, each lawmaker's ideal point $\bm{x}_u$ and each bill's
polarity $\bm{a}_d$ are vectors; the probability of a ``yes'' vote is
$\sigma(\bm{x}_u^\top \bm{a_d} +
b_d)$~\citep{heckman:1996,jackman:2001}.  When fit to data from
U.S. politics the principle dimension invariably explains most of the
variance, separating left-wing and right-wing lawmakers, and
subsequent dimensions capture other kinds of patterns in voting
behavior.  Researchers developed these models to capture the
complexity of politics beyond the left/right divide.  However, these
models are difficult to use because (as for classical factor analysis)
the dimensions are not readily interpretable---nothing ties them to
concrete issues such as \textit{Foreign Policy} or \textit{Defense}
\citep{jackman:2001}.  Our model circumvents the problem of
interpreting higher dimensions of ideal points.



The problem is that classical models only analyze the votes.  To
coherently bring issues into the picture, we need to include what the
bills are about.  Thus, the issue-adjusted model is a multidimensional
ideal point model where each additional dimension is explicitly tied
to a political issue.  The language of the bills determine which
dimensions are ``active'' when modeling the votes.  Unlike previous
multidimensional ideal point models, we do not posit higher dimensions
and then hope that they will correspond to known issues.  Rather, we
explicitly model lawmakers' votes on different issues by capturing how
the issues in a bill relate to deviations from issue-independent
voting patterns.



\subsection{Using Labeled LDA to Associate Bills with Issues}
\label{section:lda}

\begin{figure}
  \begin{center}
    \textbf{Top words in selected issues}
  \begin{tabular}{cccc}
    \hline
    \hline
    Terrorism &
    Commemorations &
    Transportation &
    Education \\
    \hline
    terrorist & nation & transportation & student \\
    September & people & minor & school \\
    attack & life & print & university \\
    nation & world & tax & charter school \\
    york & serve & land & history \\
    terrorist attack & percent & guard & nation \\
    Hezbollah & community & coast guard & child \\
    national guard & family & substitute & college \\
    \hline
  \end{tabular}
  \end{center}
  \caption{\label{figure:labeled_topics} The eight most frequent words
    from topics fit using labeled LDA~\citep{ramage:2009}.}
\end{figure}

We now describe the issue-encoding function $\bm{\theta}$.  This
function takes the language of a bill as input and returns a K-vector
that represents the proportions with which each issue is discussed.
In particular, we use labeled latent Dirichlet
allocation~\citep{ramage:2009}.  To use this method, we estimate a set
of ``topics,'' i.e., distributions over words, associated with an
existing taxonomy of political issues.  We then estimate the degree to
which each bill exhibits these topics.  This treats the text as a
noisy signal of the issues that it encodes, and we can use both tagged
bills (i.e., bills associated with a set of issues) and untagged bills
to estimate the model.

Labeled LDA is a topic model, a model that assumes that our collection
of bills can be described by a set of themes, and that each bill in
this collection is a bag-of-words drawn from a mixture of those
themes.  The themes, called topics, are distributions over a fixed
vocabulary.  In unsupervised LDA---and many other topic models---these
themes are fit to the data~\citep{blei:2003,blei:2012}.  In labeled
LDA, the themes are defined by using an existing tagging scheme.  Each
tag is associated with a topic, and its distribution is found by
taking the empirical distribution of words for documents assigned to
that tag, an approach heavily influenced by, but simpler than, that of
\cite{ramage:2009}.
 This gives interpretable names (the tags) to the topics.
(We note that our method is readily applicable to the fully
unsupervised case, i.e., for studying a political history with
untagged bills.  However, such analysis requires an additional step of
interpreting the topics.)

We used tags provided by the Congressional Research
Service~\citep{crs:2011}, a service that provides subject codes for
all bills passing through Congress.  These subject codes describe the
bills using phrases which correspond to traditional issues, such as
\emph{civil rights} and \emph{national security}.  Each bill may cover
multiple issues, so multiple codes may apply to each bill. (Many bills
have more than twenty labels.)  Figure~\ref{figure:labeled_topics}
illustrates the top words from several of these labeled topics.  We
then performed two iterations of unsupervised LDA (\citep{blei:2003}
with variational inference to smooth the word counts in these topics.
We used the 74 issues in all (the most-frequent issue labels); we
summarize all 74 of them in \myappendix{topics}.

With topics in hand, we model each bill with a mixed-membership model:
Each bill is drawn from a mixture of the topics, but each one exhibits
them with different proportions.  Denote the $K$ topics by
$\beta_{1:K}$ and let $\alpha$ be a vector of Dirichlet parameters.
The generative process for each bill $d$ is:
\begin{enumerate}
\item Choose topic proportions $\theta_d \sim \textrm{Dirichlet}(\alpha)$.
\item For each word $n \in \{1, \ldots, N\}$:
  \begin{enumerate}
  \item Choose a topic assignment $z_{d,n} \sim \theta_d$.
  \item Choose a word $w_{d,n} \sim \beta_{z_{d,n}}$.
  \end{enumerate}
\end{enumerate}
The function $\theta(\bm{w}_d)$ is the posterior expectation of
$\theta_d$.  It represents the degree to which the bill exhibits the
$K$ topics, where those topics are explicitly tied to political issues
through the congressional codes, and it is estimated using variational
inference at the document level~\citep{blei:2003}.
The topic modeling portion of the model is illustrated on the left
hand side of the graphical model in \myfig{legis_gm}.

We have completed our specification of the model.  Given roll call
data and bill texts, we first compute the issue vectors for each bill.
We then use these in the issue-adjusted ideal point model of
\myfig{legis_gm} to infer each legislator's posterior ideal point and
per-issue adjustment.  We now turn to the central computational
problem for this model, posterior inference.


\section{Posterior Estimation}
\label{section:inference}
Given roll call data and an encoding of the bills to issues, we form
inferences and predictions through the posterior distribution of the
latent ideal points, issue adjustments, and bill variables, $p(x, \bm
z, a, b | \bm{v}, \bm{\theta})$. In the next section, we inspect this
posterior to explore lawmakers' positions about specific issues.

As for most interesting Bayesian models, this posterior is not
tractable to compute; we must approximate it.  Approximate posterior
inference for Bayesian ideal point models is usually performed with
MCMC methods, such as Gibbs
sampling~\citep{johnson:1999ch6,jackman:2001,martin:2002,clinton:2004}.
Here we will develop an alternative algorithm based on variational
inference.  Variational inference tends to be faster than MCMC, can
handle larger data sets, and is attractive when fast Gibbs updates are
not available.  In the next section, we will use variational inference
to analyze twelve years of roll call data.

\subsection{Mean-field Variational Inference}

In variational inference we select a simplified family of candidate
distributions over the latent variables and then find the member of
that family which is closest in KL divergence to the posterior of
interest~\citep{jordan:1999,Wainwright:2008}.  This turns the problem
of posterior inference into an optimization problem.  For posterior
inference in the issue-adjusted model, we use the fully-factorized
family of distributions over the latent variables, i.e., the
mean-field family,
\begin{align}
q(x, & \bm y, \bm z, a, b | \bm \eta ) =
\left( \prod_U \mathcal{N}(x_u | \tilde x_u, \sigma_{x}^2 )
  \mathcal{N}(\bm z_u | \bm \tilde z_u, \sigma_{z}^2 ) \right)
  \left( \prod_D \mathcal{N}(a_d | \tilde a_d, \sigma_{a}^2)
  \mathcal{N}(b_d | \tilde b_d, \sigma_{b}^2) \right).
  \label{equation:variational_posterior}
\end{align}
This family is indexed by the \textit{variational parameters} $\bm
\eta = \big\{ (\tilde x_u, \sigma_{x})$, $(\bm \tilde z_u, \sigma_{\bm
  z_u})$, $(\tilde a, \sigma_{a})$, $(\tilde b, \sigma_{b}) \big\}$,
which specify the means and variances of the random variables in the
variational posterior.  While the model specifies priors over the
latent variables, in the variational family each instance of each
latent variable, such as each lawmaker's issue adjustment for
\textit{Taxation}, is endowed with its own variational distribution.
This lets us capture data-specific marginals---for example, that one
lawmaker is more conservative about \textit{Taxation} while another is
more liberal.

We fit the variational parameters to minimize the KL divergence
between the variational posterior and the true posterior. Once fit, we
can use the variational means to form predictions and posterior
descriptive statistics of the lawmakers' issue adjustments.  In ideal
point models, the means of a variational distribution can be excellent
proxies for those of the true posterior~\citep{gerrish:2011}).

\subsection{The Variational Objective}

Variational inference proceeds by taking the fully-factorized
distribution (\myeq{variational_posterior}) and successively updating
the parameters $\bm \eta$ to minimize the KL divergence between the
variational distribution (Equation~\ref{equation:variational_posterior}) and
the true posterior:
\begin{align}
  \label{equation:kl_divergence}
  \hat \eta = \arg \min_\eta \mbox{KL}\left( q_\eta(x, \bm z, a, b) ||
  p(x, \bm a, a, b | v) \right)
\end{align}
This optimization is usually reformulated as the problem of maximizing
a lower bound (found via Jensen's inequality) on the marginal
probability of the observations:
\begin{align}
  \label{equation:elbo}
  p(v) = & \int_{\bm \eta} p(x, \bm z, a, b, v) dx d{\bm z} da db \nonumber \\
  \ge & \int_{\bm \eta} q_\eta(x, \bm z, a, b) \log \frac{p(x, z, a, b, v)}{q_\eta(x, \bm z, a, b)} dx d \bm z da db \nonumber \\
  = & \expectq{p(x, \bm z, a, b, v)} - \expectq{q_\eta(x, \bm z, a, b)} = \mathcal{L}_\eta.
\end{align}
We follow the example of \cite{braun:2007} by referring to the lower
bound $\mathcal{L}_\eta$ as the \emph{evidence lower bound} (ELBO).

For many models, the ELBO can be expanded as a closed-form function of
the variational parameters and then optimized with gradient ascent or
coordinate ascent.  However, the issue-adjusted ideal point model does
not allow for a closed-form objective.  Previous research on such
non-conjugate models overcomes this by approximating the ELBO
\citep{braun:2007,gerrish:2011}.  Such methods are effective, but they
require many model-specific algebraic tricks and tedious derivations.
Here we take an alternative approach, where we approximate the
gradient of the ELBO with Monte-Carlo integration and perform
stochastic gradient ascent with this approximation.  This gave us an
easier way to fit the variational objective for our complex model.

\subsection{Optimizing the Variational Objective with a Stochastic
  Gradient}
We begin by computing the gradient of the ELBO in \myeq{elbo}.  We
rewrite it in terms of integrals, then exchange the order of
integration and differentiation,
and apply the chain rule:
\begin{align}
     \label{equation:gradient}
 \nabla \mathcal{L}_\eta &= \nabla \Bigl[
  \int q_{\eta}(x, \bm z, a, b) (\log p(x, \bm z, a, b, v) - \log q_{\eta}(x, \bm z, a, b))
  dx \Bigr] \\ \nonumber
  & = \int \nabla \Bigl[ q_{\eta}(x, \bm z, a, b) (\log p(x, \bm z, a, b, v) - \log
     q_{\eta}(x, \bm z, a, b)) \Bigr] dx  \\ \nonumber
     &= \int \nabla q_{\eta}(x, \bm z, a, b) (
     \log p(x, \bm z, a, b, v) - \log q_{\eta}(x, \bm z, a, b)) - q_{\eta}(x, \bm z, a, b) \nabla \log q_{\eta} dx. 
\end{align}
Above we have assumed that the support of $q_\eta$ is not a function
of $\eta$, and that $\log q_\eta(x, \bm z, a, b)$ and $\nabla \log
q_\eta(x, \bm z, a, b)$ are continuous with respect to $\eta$.

We can rewrite \myeq{gradient} as an expectation by using the
identity $q_\eta(x) \nabla \log q_\eta(x) = \nabla q_\eta(x)$:
\begin{align}
  \nabla \mathcal{L}_\eta =
  \expectq{\nabla \log q_\eta(x, \bm z, a, b) \left(\log p(x, \bm z, a, b, v) - \log
      q_{\eta}(x, \bm z, a, b) - 1 \right)}.
  \label{equation:gradient-as-expectation}
\end{align}
Next we use Monte Carlo integration to form an unbiased estimate of the
gradient at $\eta = \eta_0$.  We obtain $M$ \emph{iid} samples $(x_{1},
\ldots, x_{M}, \ldots, b_1, \ldots, b_M$) from the variational distribution $q_{\eta_0}$
for the approximation
\begin{align}
  \nabla \mathcal{L}_\eta \Bigr|_{\eta_0} & \approx \nonumber \\
  & \hspace{-25pt} \frac{1}{M} \sum_{m=1}^M
  \nabla \log q_{\eta}(x_{m}, \bm z_m, a_m, b_m) \Bigr|_{\eta_0}
  (\log p(x_{m}, \bm z_m, a_m, b_m, y) - \log q_{\eta_0}(x_{m}, \bm z_m, a_m, b_m) - C).
  \label{equation:svo_gradient}
\end{align}
We denote this approximation $\tilde{\nabla} {\cal L}_\eta
|_{\eta_0}$. Note we replaced the 1 in \myeq{gradient-as-expectation}
with a constant $C$, which does not affect the expected value of the
gradient (this follows because $\expectq{\nabla \log q_\eta(x, \bm z,
  a, b)} = 0$).  We discuss in the supplementary materials how to set
$C$ to minimize variance.  Related estimates of similar gradients have
been studied in recent
work~\citep{carbonetto:2009,graves:2011,paisley:2012} and in the
context of expectation maximization~\citep{wei:1990}.

Using this method for finding an approximate gradient, we optimize the
ELBO with stochastic
optimization~\citep{robbins:1951,Spall:2003,bottou:2004}.  Stochastic
optimization follows noisy estimates of the gradient with a decreasing
step-size.  While stochastic optimization alone is sufficient to
achieve convergence, it may take a long time to converge.  To improve
convergence rates, we used two additional ideas: quasi-Monte Carlo
samples (which minimize variance) and second-order updates (which
eliminate the need to select an optimization parameter).  We provide
details of these improvements in the appendix.

Let us return briefly to the problem that motivated this section.  Our
goal is to estimate the mean of the hidden random variables---such as
lawmakers' issue adjustments $\bm z$---from their votes on bills.  We
achieved this by variational Bayes, which amounts to maximizing the
ELBO (\myeq{elbo}) with respect to the variational parameters.  This
maximization is achieved with stochastic optimization on
\myeq{svo_gradient}.  In the next section we will empirically study
these inferred variables (i.e., the expectations induced by the
variational distribution) to better understand distinctive voting
behavior.


\section{Issue adjustments in the United States Congress}
\label{section:empirical_analysis}

We used the issue-adjusted ideal point model to study the complete
roll call record from the United States Senate and House of
Representatives during the years 1999-2010.  We report on this study
in this and the next section.  We first evaluate the model fitness to
this data, confirming that issue-adjustments give a better model of
roll call data and that the encoding of bills to issues is responsible
for the improvement.  We then use our inferences to give a qualitative
look at U.S. lawmakers' issue preferences, demonstrating how to use
our richer model of lawmaker behavior to explore a political history.

\subsection{The United States Congress from 1999-2010}
We studied U.S. Senate and House of Representative roll-call votes
from 1999 to 2010.  This period spanned Congresses 106 to 111, the
majority of which Republican President George W. Bush held office.
Bush's inauguration and the attacks of September 11th, 2001 marked the
first quarter of this period, followed by the wars in Iraq and
Afghanistan.  Democrats gained a significant share of seats from 2007
to 2010, taking the majority from Republicans in both the House and
the Senate. Democratic President Barack Obama was inaugurated in
January 2009.

The roll-call votes are recorded when at least one lawmaker wants an
explicit record of the votes on the bill.  For a lawmaker, such
records are useful to demonstrate his or her positions on issues.
Roll calls serve as an incontrovertible record for any lawmaker who
wants one.  We downloaded both roll-call tables and bills from
\verb!www.govtrack.us!, a nonpartisan website which provides records
of U.S. Congressional voting.  Not all bill texts were available, and
we ignored votes on bills that did not receive a roll call, but we had
over one hundred for each Congress.  Table~\ref{fig:data_stats}
summarizes the statistics of our data.

\begin{figure}
  \center
  \caption{Roll-call data sets used in the experiments.  These counts
    include votes in both the House and Senate.  Congress 107 had
    fewer votes than the remaining congresses in part because this
    period included large shifts in party power, in addition to the
    attacks on September 11th, 2001.  The number of lawmakers within
    each House and Senate varies by congress because there was some
    turnover within each Congress.  In addition, some lawmakers never
    voted on legislation in our experiments (recall, we used
    legislation for which both text was available and for which the
    roll-call was recorded).}
  \begin{small}
  \center{\textbf{Statistics for the U.S. Senate}} \\
  \begin{tabular}{ccccc}
    \hline
    \hline
    \hspace{-8pt} \textbf{Congress} \hspace{-8pt} & \hspace{-8pt} {\textbf{Years}} \hspace{-8pt} & \textbf{Lawmakers} \hspace{-8pt} & \hspace{-4pt} \textbf{Bills} \hspace{-4pt} & \hspace{-4pt} \textbf{Votes} \\
    \hline
    106 & 1999-2000 & 81 & 101 & 7,612 \\
    107 & 2001-2002 & 78 & 76 & 5,547 \\
    108 & 2003-2004 & 101 & 83 & 7,830 \\
    109 & 2005-2006 & 102 & 74 & 7,071 \\
    110 & 2007-2008 & 103 & 97 & 9,019 \\
    111 & 2009-2010 & 110 & 62 & 5,936 \\
    \hline
  \end{tabular}
  \\
    \center{\textbf{Statistics for the U.S. House of Representatives}} \\
  \begin{tabular}{ccccc}
    \hline
    \hline
    \hspace{-8pt} \textbf{Congress} \hspace{-8pt} & \hspace{-8pt} {\textbf{Years}} \hspace{-8pt} & \textbf{Lawmakers} \hspace{-8pt} & \hspace{-4pt} \textbf{Bills} \hspace{-4pt} & \hspace{-4pt} \textbf{Votes} \\
    \hline
    106 & 1999-2000 & 437 & 345 & 142,623 \\
    107 & 2001-2002 & 61 & 360 & 18,449 \\
    108 & 2003-2004 & 440 & 490 & 200,154 \\
    109 & 2005-2006 & 441 & 458 & 187,067 \\
    110 & 2007-2008 & 449 & 705 & 287,645 \\
    111 & 2009-2010 & 446 & 810 & 330,956 \\
    \hline
  \end{tabular}
  \end{small}
  \label{fig:data_stats}
\end{figure}


We fit our models to two-year periods in the House and (separately) to
two-year periods in the Senate.  Some bills received votes in both the
House and Senate; in those cases, the issue-adjusted model's treatment
of the bill in the House was completely independent of its treatment
by the model in the Senate.

\paragraph{Vocabulary.}
To fit the labeled topic model to each bill, we represented each bill
as a vector of phrase counts.  This ``bag of phrases'' is similar to
the ``bag of words'' assumption commonly used in natural language
processing.  To select this vocabulary, we considered all phrases of
length one word to five words.  We then omitted content-free phrases
such as ``and'', ``when'', and ``to the''.  The full vocabulary
consisted of 5,000 $n$-grams (further details of vocabulary selection
are in \myappendix{vocabulary}).  We used these phrases to
algorithmically define topics and assign issue weights to bills as
described in \mysec{lda}.


\paragraph{Identification.}  When using ideal-point models for
interpretation, we must address the issue of identification.  The
signs of ideal points $x_u$ and bill polarities $a_d$ are arbitrary,
for example, because $x_u a_d = (-x_u)(-a_d)$. This leads to a
multimodal posterior \citep{jackman:2001}.  We address this by
flipping ideal points and bill polarities if necessary to follow the
convention that Republicans are generally on the right (positive on
the line) and Democrats are generally on the left (negative on the
line).

\subsection{Ideal Point Models vs. Issue-adjusted Ideal Point Models}
\label{section:jackman_vs_exploratory}

The issue-adjusted ideal point model in Equation~\ref{equation:ia-ipm}
is a generalization of the traditional ideal point model (see
\mysec{relationship}).  Before using this more complicated model to
explore our data, we empirically justify this increased complexity.
We first outline empirical differences between issue-adjusted ideal
points and traditional idea points.  We then report on a quantitative
validation of the issue-adjusted model.

\paragraph{Examples: adjusting for issues.}
To give a sense of how the issue-adjusted ideal point model works,
\mytable{issue_adjustments} gives a side-by-side comparison of
traditional ideal points $x_u$ and issue-adjusted ideal points $(x_u +
\bm z_u^T \bm \theta)$ for the ten most-improved bills of Congress 111
(2009-2010).  For each bill, the top row shows the ideal points of
lawmakers who voted ``Yea'' on the bill and the bottom row shows
lawmakers who voted ``Nay''.  The top and bottom rows are a partition
of votes rather than separate treatments of the same votes.  In a good
model of roll call data, these two sets of points will be separated,
and the model can place the bill parameters at the correct cut point.
Over the whole data set, the cut point of the votes improved in 14,347
heldout votes.  (It got worse in 8,304 votes and stayed the same in
5.7M.)


\begin{figure}
  \center
  \rowcolors{1}{gray!30}{}
  \setlength{\extrarowheight}{1.5pt}
  \renewcommand{\arraystretch}{1.5}
  \begin{tabular}{|p{4.3cm}|c|c|}
\hline
\small \textbf{Bill description}
& \small \textbf{Votes by ideal point}
& \small \textbf{Votes by adjusted point} \\
\hline
\small
H. Res 806 (amending an education/environment trust fund)
 & \includegraphics[width=0.35\textwidth]{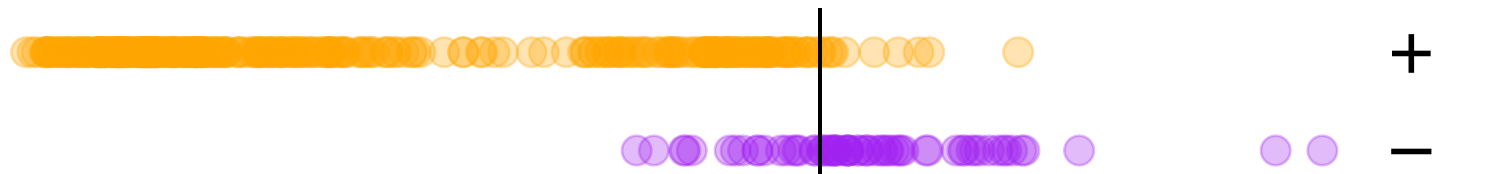}
 & \includegraphics[width=0.35\textwidth]{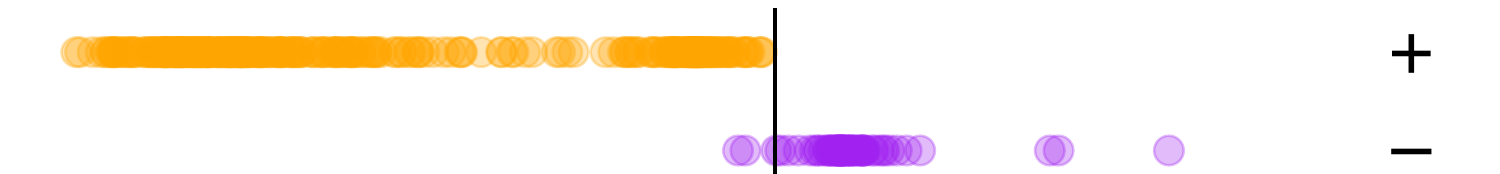} \\

Providing for conditional adjournment/recess of Congress
& \includegraphics[width=0.35\textwidth]{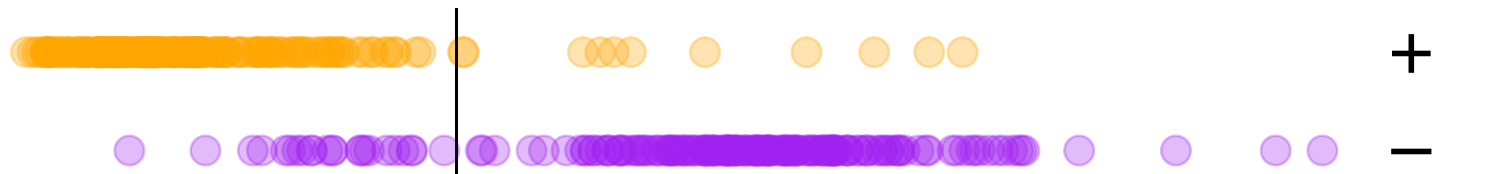}
& \includegraphics[width=0.35\textwidth]{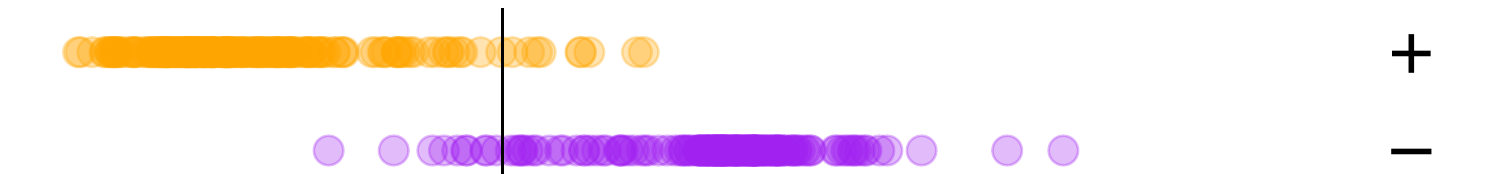} \\

Establish R\&D program for gas turbines 
& \includegraphics[width=0.35\textwidth]{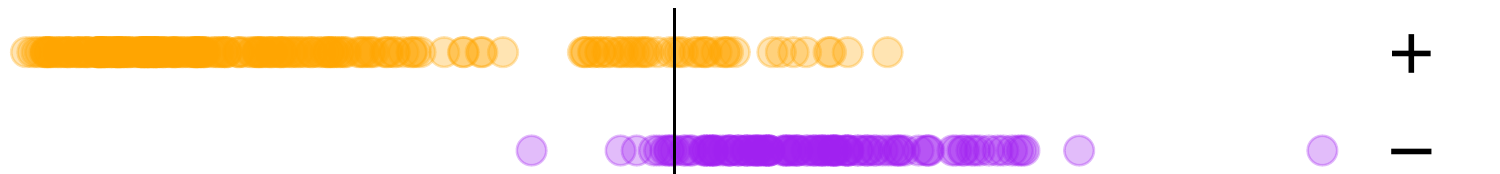}
& \includegraphics[width=0.35\textwidth]{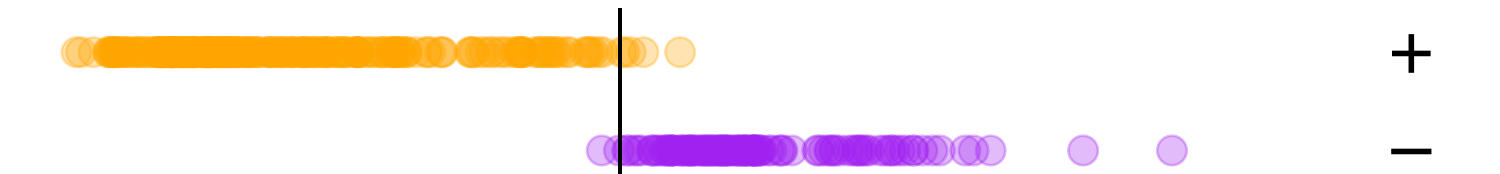} \\

Recognizing AmeriCorps and community service
& \includegraphics[width=0.35\textwidth]{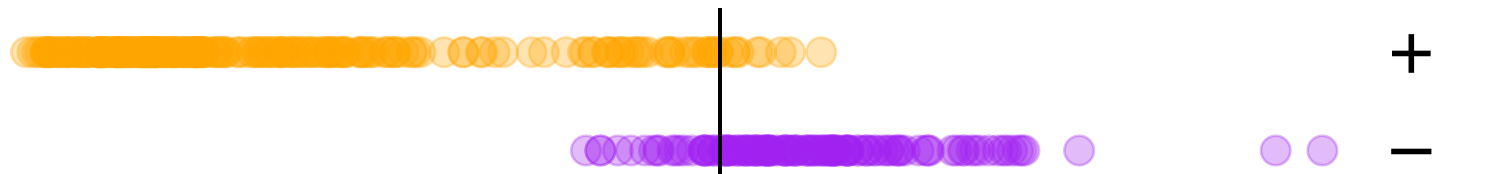}
& \includegraphics[width=0.35\textwidth]{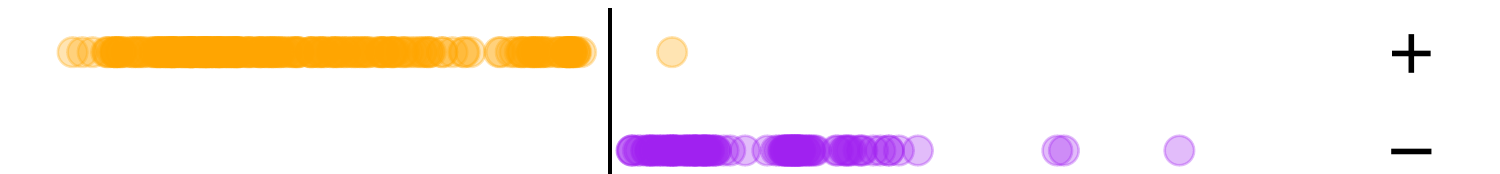} \\

Providing for conditional adjournment of Congress
& \includegraphics[width=0.35\textwidth]{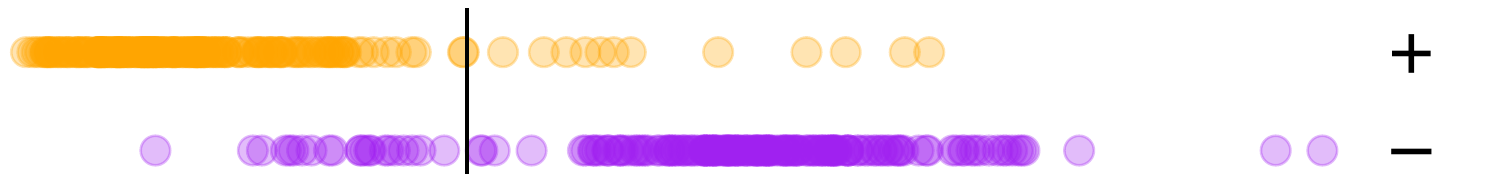}
& \includegraphics[width=0.35\textwidth]{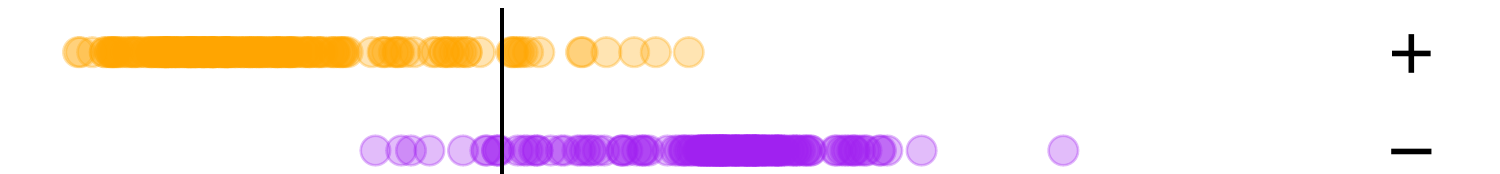} \\

Providing for the sine die adjournment of Congress
& \includegraphics[width=0.35\textwidth]{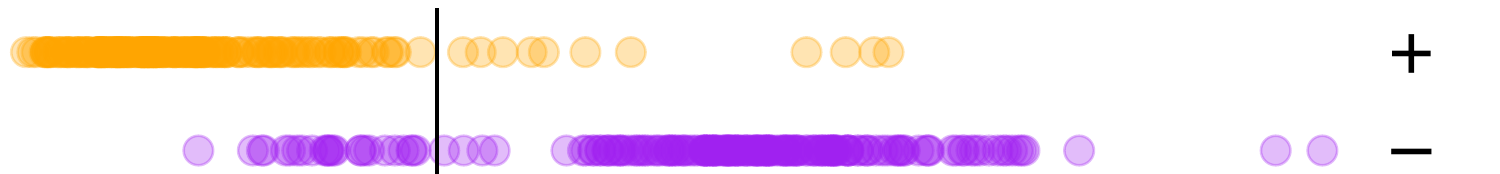}
& \includegraphics[width=0.35\textwidth]{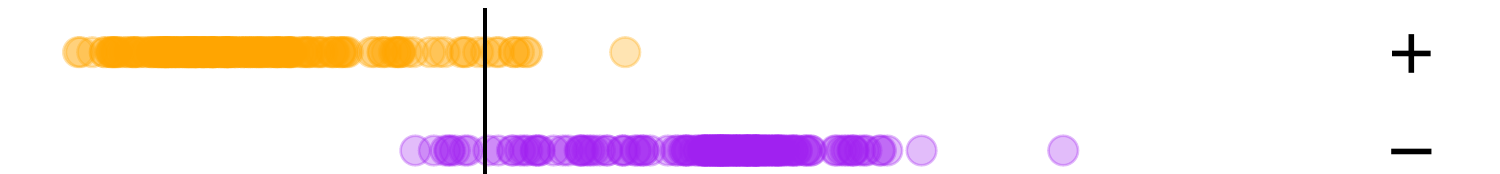} \\

Providing for an adjournment / recess of Congress
& \includegraphics[width=0.35\textwidth]{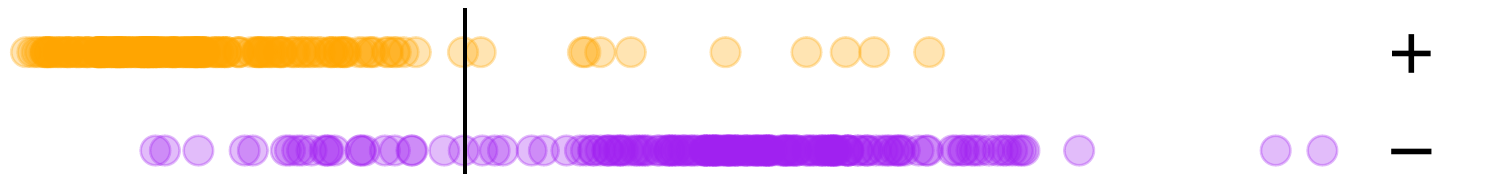}
& \includegraphics[width=0.35\textwidth]{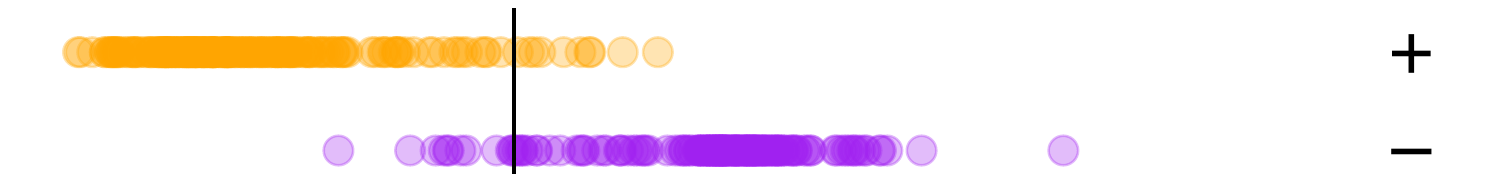} \\

Preventing child marriage in developing countries
& \includegraphics[width=0.35\textwidth]{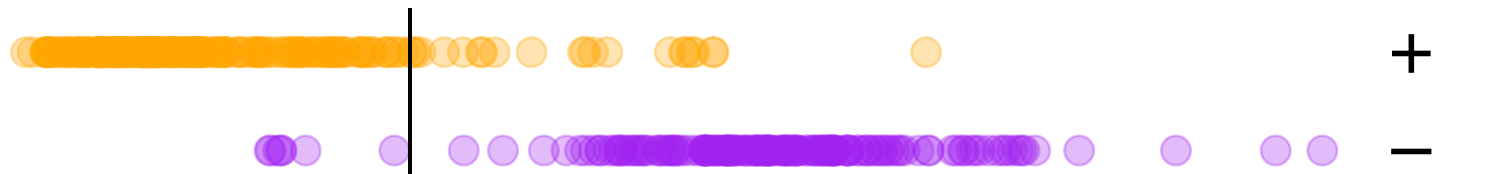}
& \includegraphics[width=0.35\textwidth]{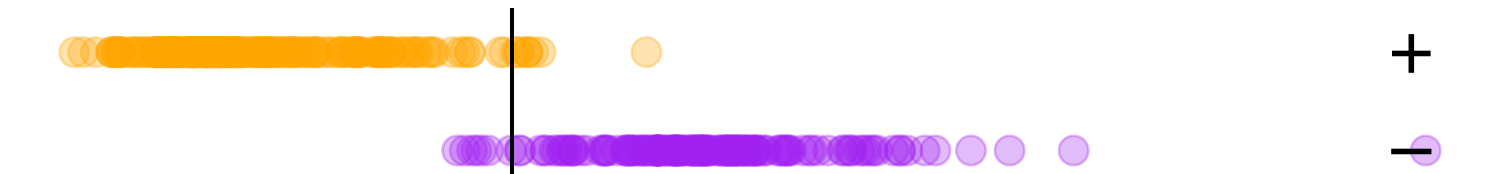} \\

Providing for a conditional House adjournment
& \includegraphics[width=0.35\textwidth]{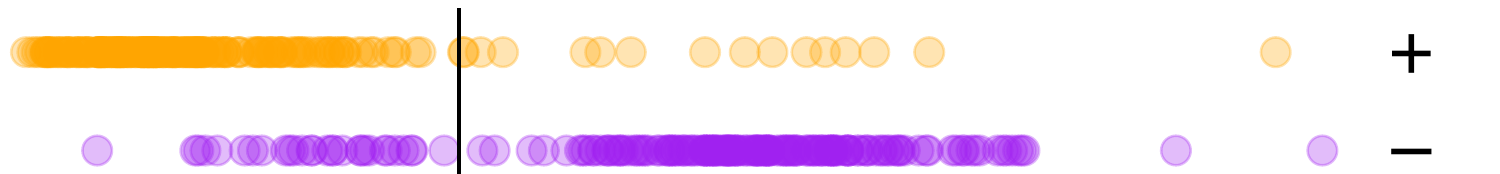}
& \includegraphics[width=0.35\textwidth]{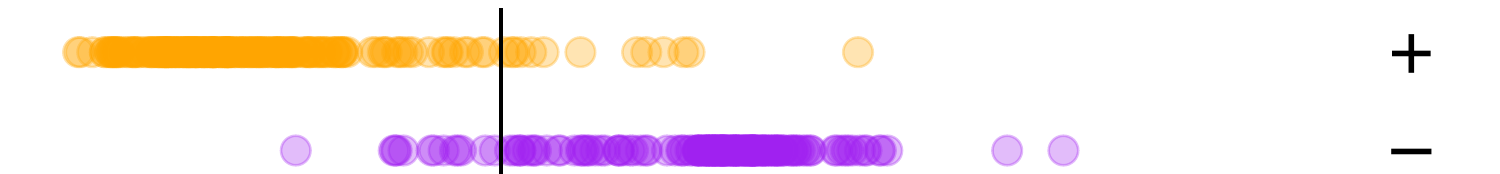} \\

Congratulating UMD Men's basketball
& \includegraphics[width=0.35\textwidth]{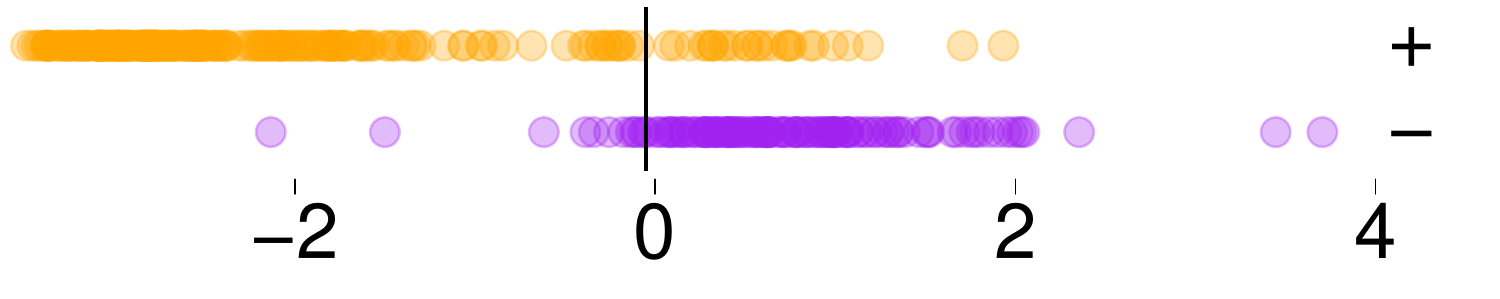}
& \includegraphics[width=0.35\textwidth]{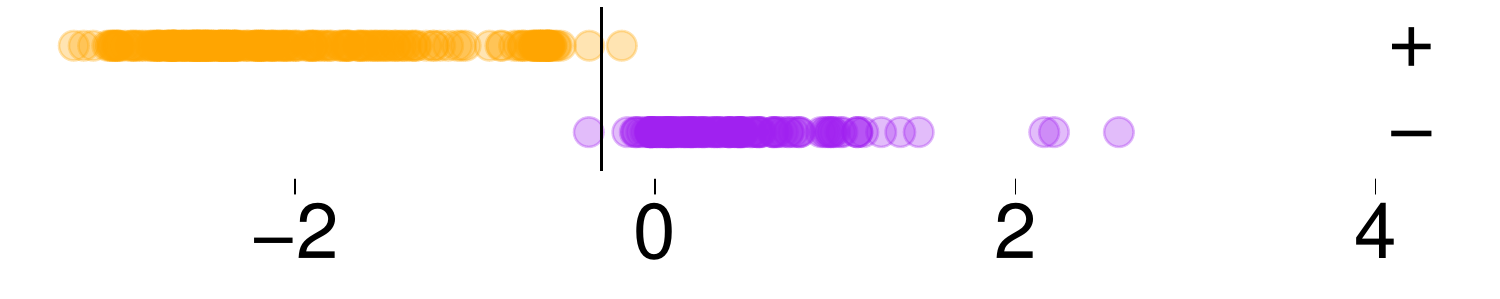} \\
\hline
  \end{tabular}
  \caption{Issue-adjusted ideal points can explain votes better than
  standard ideal points.  The x-axis of each small plot shows ideal
  point or issue-adjusted ideal point for a lawmaker. Each bill's
  indifference point $- b_d / a_d$ is shown as a vertical
  line. Positive votes (orange) and negative votes (purple) are
  better-divided by issue-adjusted ideal points.}
  \label{table:issue_adjustments}
\end{figure}

\paragraph{Comparing issue-adjusted ideal points to traditional ideal
  points.}
The traditional ideal point model (\myeq{trad_ipm}) uses one variable
per lawmaker, the ideal point $x_u$, to explain all of her voting
behavior.  In contrast, the issue-adjusted model
(Equation~\ref{equation:ia-ipm}) uses $x_u$ along with $K$ issue
adjustments.  Here we ask, how does does $x_u$ under these two models
differ?  We fit ideal points to the 111th House (2009 to 2010) and
issue-adjusted ideal points to the same period with regularization
$\lambda=1$.

\begin{figure}
  \center
  \includegraphics[width=0.8\textwidth]{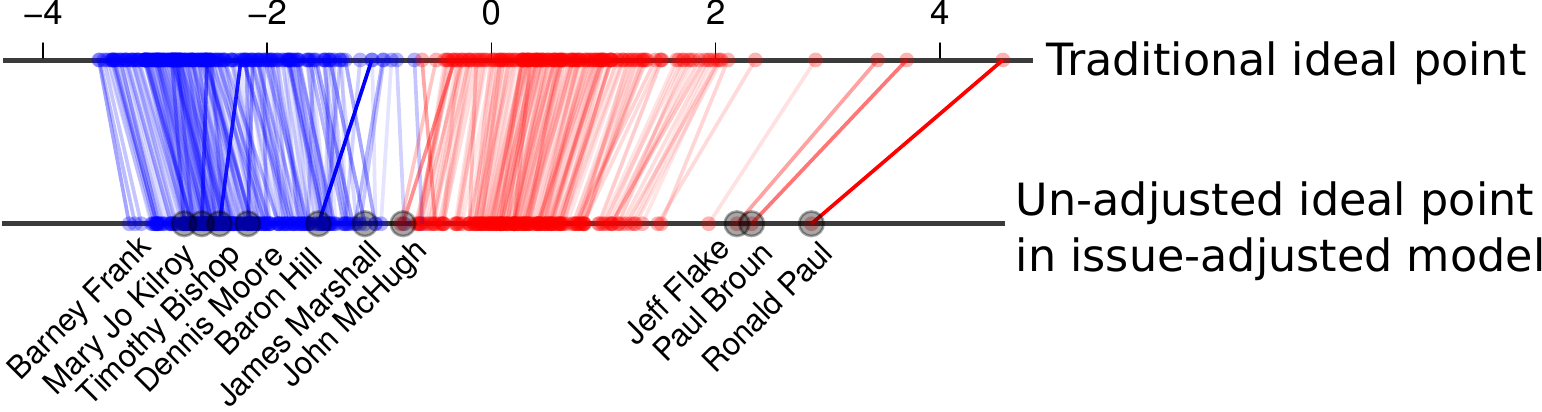} \\
  \includegraphics[width=0.8\textwidth]{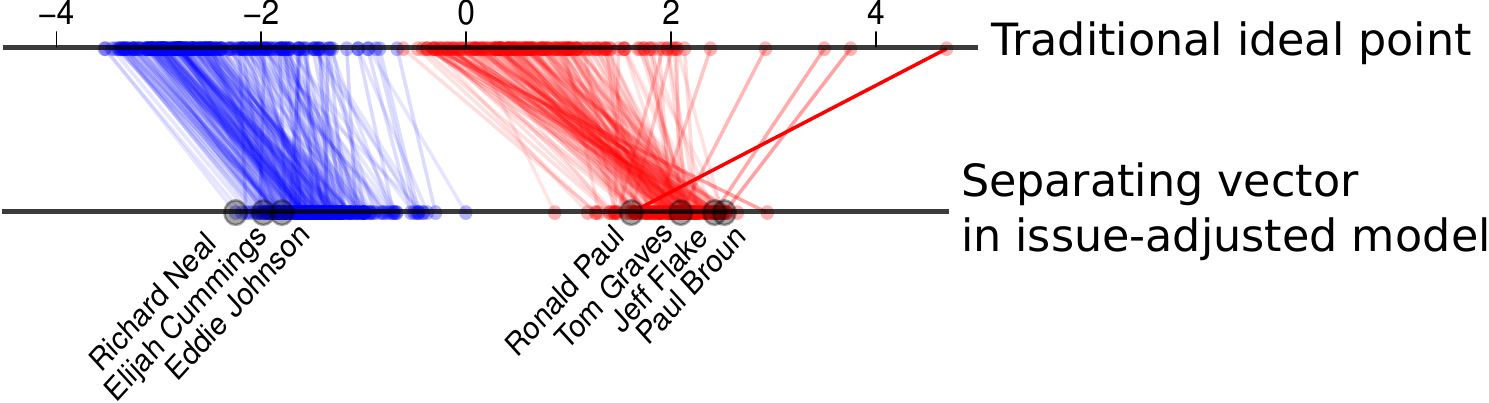}
  \label{figure:jackman_vs_offset}
  \caption{Classic issue-adjusted ideal points $x_u$ (top row, both
    figures) separate lawmakers by party better than un-adjusted ideal
    points $x_u$ from the issue-adjusted model (bottom row, top
    figure).  The issue-adjusted model can still separate Republicans
    from Democrats better than the ideal point model along a
    separating vector (bottom row, bottom figure).  In each figure,
    Republicans are colored red, and Democrats are blue.  These ideal
    points were estimated in the 111th House of Representatives.  The
    line connecting ideal points from each model has opacity
    proportional to the squared residuals in a linear model fit to
    predict issue-adjusted ideal points from ideal points.
    The separating vector was defined using linear discriminant analysis.}
\end{figure}

The top panel of \myfig{jackman_vs_offset} compares the classical ideal
points to the global ideal points from the issue-adjusted
model.
In this parallel plot, the top axis of this represents a lawmaker's
ideal point $x_u$ under the classical model, while the bottom axis
represents his global ideal point under the issue-adjusted model.  (We
will use plots like this again in this paper.  It is called a parallel
plot, and it compares separate treatments of lawmakers.  Lines between
the same lawmakers under different treatment are shaded based on their
deviation from a linear model to highlight unique lawmakers.) The
ideal points in \myfig{jackman_vs_offset} are similar; their
correlation coefficient is 0.998. The most noteworthy difference is
that lawmakers appear more partisan under the traditional ideal point
model---enough that Democrats are completely separated from
Republicans by $x_u$---while issue-adjusted ideal points provide a
softer split.

This is not surprising, because the issue-adjusted model is able to
use lawmakers' adjustments to explain their votes.  In fact, the
political parties are \emph{better} separated with issue adjustments
than they are by ideal points alone.  We checked this by writing each
lawmaker $u$ as the vector $w_u := (x_u, z_{u,1}, \ldots, z_{u,k})$
and performing linear discriminant analysis to find that vector
$\beta$ which ``best'' separates lawmakers by party along $w_u^T
\beta$.

We illustrate lawmakers' projections $w_u^T \beta$ along the
discriminant vector $\beta$ in the bottom figure of
\myfig{jackman_vs_offset} (we normalized variance of these projections
to match that of the ideal points).  The correlation coefficient
between this prediction and political party is $0.979$, much higher
than the correlation between ideal points $x_u$ and political party
($0.921$).

To be sure, some of this can be explained by random variation in the
additional 74 dimensions.  To check the extent of this improvement due
only to dimension, we draw random issue adjustments from normal random
variables with the same variance as the empirically observed issue
adjustments.  In 100 tests like this, the correlation coefficient was
higher than for classical ideal points, but not by much: $0.933 \pm
0.004$.  Thus, the posterior issue adjustments provide a signal for
separating the political parties \emph{better} than ideal points
alone.  In fact, we will see in \mysec{procedural_cartel_theory} that
procedural votes driven by political ideology is one of the factors
driving this improvement.


\begin{figure}
  \begin{tabular}{|m{0.85in}|m{2.54in}|m{2.54in}|}
    \hline
    & Popularity & Polarity \\
    \hline
    Not procedural &
    \includegraphics[width=0.4\textwidth]{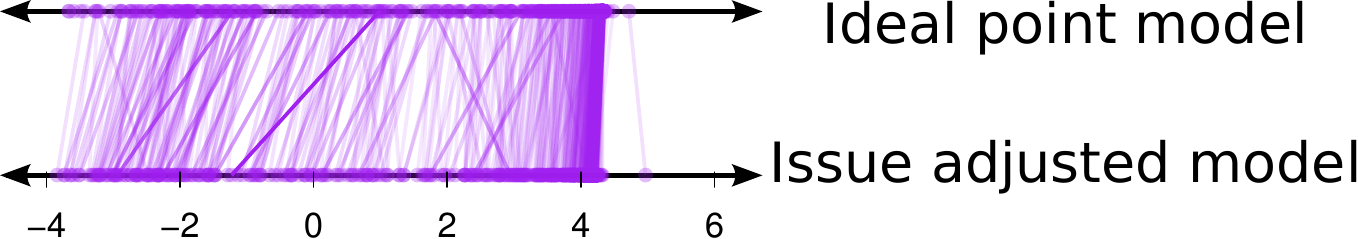} &
    \includegraphics[width=0.4\textwidth]{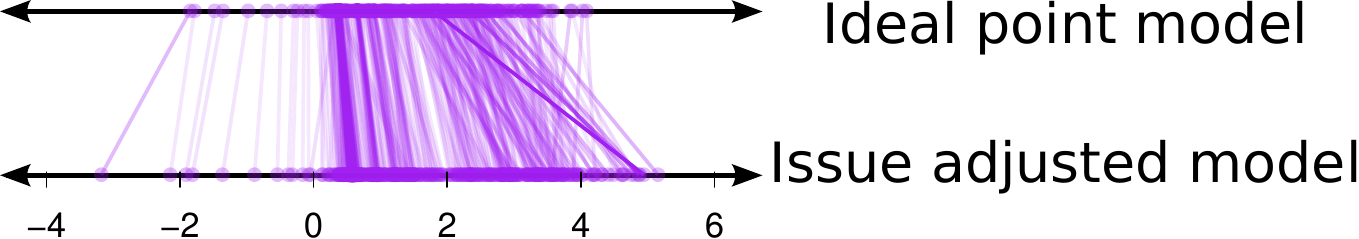} \\
    Procedural &
    \includegraphics[width=0.4\textwidth]{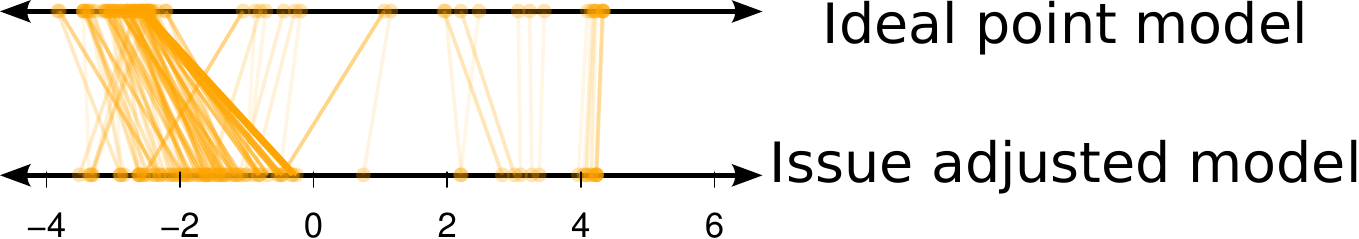} &
    \includegraphics[width=0.4\textwidth]{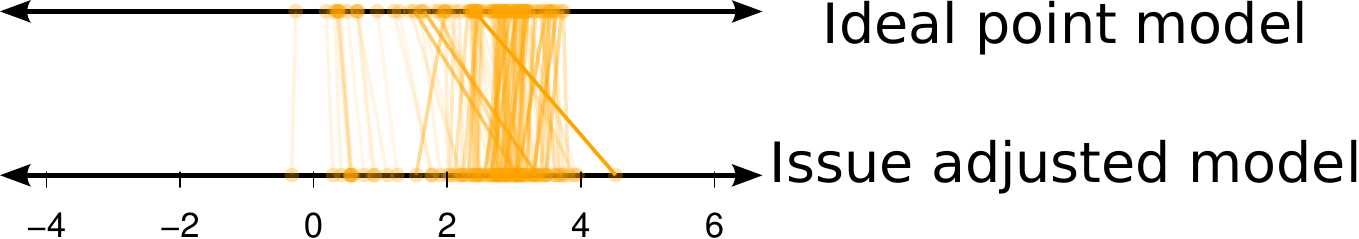} \\
    \hline
  \end{tabular}
  \caption{Procedural bills are more popular under the issue-adjusted
    voting model. Top: popularity $b_d$ of procedural bills under the
    issue-adjusted voting model is greater than with traditional ideal
    points.  Bottom: consistent with \citet{cox:2002} and
    \emph{procedural cartel theory}, the polarity of procedural bills
    is generally more extreme than that of non-procedural bills.
    However, issue adjustments lead to increased polarity (i.e.,
    certainty) among non-procedural votes as well.  The procedural
    issues include \emph{congressional reporting requirements},
    \emph{government operations and politics}, \emph{House of
      Representatives}, \emph{House rules and procedure},
    \emph{legislative rules and procedure}, and \emph{Congress}.}
  \label{figure:bills_parameter_changes}
\end{figure}

\paragraph{Changes in bills' parameters.}
Bills' polarity $a_d$ and popularity $b_d$ are similar under both the
traditional ideal point model and the issue-adjusted model. We
illustrate bills' parameters in these two models in
\myfig{bills_parameter_changes} and note some exceptions.

First, procedural bills stand out from other bills in becoming more
popular overall.  In \myfig{bills_parameter_changes}, procedural bills
have been separated from traditional ideal points.  We attribute the
difference in procedural bills' parameters to \emph{procedural cartel
  theory}, which we describe further in
\mysec{procedural_cartel_theory}.

The remaining bills have also become less popular but more polarized
under the issue-adjusted model.  This is because the issue-adjusted
model represents the interaction between lawmakers and bills with $K$
additional bill-specific variables, all of which are mediated by the
bill's polarity.  This means that the the model is able to depend more
on bills' polarities than bills' popularities to explain votes.  For
example, Donald Young regularly voted \emph{against} honorary names
for regional post offices.  These bills---usually very popular---would
have high popularity under the ideal point model.  The issue-adjusted
model also assigns high popularity to these bills, but it takes
advantage of lawmaker's positions on the \emph{postal facilities}
issue to explain votes, decreasing reliance on the bill's popularity
(\emph{postal facilities} was more common than 50\% of other issues,
including \emph{human rights}, \emph{finance}, and \emph{terrorism}).

\subsection{Evaluation of the Predictive Distribution}
\label{section:performance}

We have described the qualitative differences between the
issue-adjusted model and the traditional ideal point model.  We now
turn to a quantitative evaluation: Does the issue-adjusted model give
a better fit to legislative data?

We answer this question via cross validation and the predictive
distribution of votes.  For each session, we divide the votes, i.e.,
individual lawmaker/bill pairs, into folds.  For each fold, we hold
out the votes assigned to it, fit our models to the remaining votes,
and then evaluate the log probability of the held out votes under the
predictive distribution.  A better model will assign higher
probability to the held-out data.  We compared several methods:
\begin{enumerate}
\item The issue-adjusted ideal point model with topics found by
  labeled LDA: This is the model and algorithm described above.  We
  used a regularization parameter $\lambda = 1$.  (See
  \myappendix{hyperparameters} for a study of the effect of
  regularization.)


\item The issue-adjusted ideal point model with explicit labels on the
  bills: Rather than infer topics with labeled LDA, we used the CRS
  labels explicitly.  If a bill contains $J$ labels, we gave it weight
  $1/J$ at each of the corresponding components of the topic vector
  $\theta$.
\item The traditional ideal point model of \cite{clinton:2004}: This
  model makes no reference to issues.  To manage the scale of the
  data, and keep the comparison fair, we used variational inference.
  (In \cite{gerrish:2011}, we showed that variational approximations
  find as good approximate posteriors as MCMC in ideal point models.)
\item A permuted issue-adjusted model: Here, we selected a random
  permutation $\pi \in S_D,$ to shuffle the $D$ topic vectors
  $\bm \theta_d \rightarrow \bm \theta_{\pi(d)}$
  and fit the issue-adjusted model with the permuted vectors.  This
  permutation test removes the information contained in matching bills
  to issues, though it maintains the same empirical distribution over
  topic mixtures.  It can indicate that improvement we see over
  traditional ideal points is due to the bills' topics, not due to
  spurious factors (such as the change in dimension).  In this method
  we used five random permutations.
\end{enumerate}

\begin{figure}
  \caption{Average log-likelihood of heldout votes across all sessions
    for the House and Senate.  Log-likelihood was averaged across
    folds using six-fold cross validation for Congresses 106 to 111
    (1999-2010) with regularization $\lambda=1$.  The variational
    distribution had higher heldout log-likelihood for all congresses
    in both chambers than either } \center \small
  \center
  \textbf{Heldout log likelihood of Senate votes} \\
  \begin{tabular}{ccccccc}
    \hline
    \hline
    \textbf{Congress} & \hspace{-4pt} 106 \hspace{-5pt}
    & \hspace{-4pt} 107 \hspace{-5pt}
    & \hspace{-4pt} 108 \hspace{-5pt}
    & \hspace{-4pt} 109 \hspace{-5pt}
    & \hspace{-4pt} 110 \hspace{-5pt}
    & \hspace{-4pt} 111 \hspace{-4pt} \\
    \hline
    Traditional ideal point model (IPM)
    & \hspace{-4pt} -0.209 \hspace{-5pt}
    & \hspace{-4pt} \textbf{-0.209} \hspace{-5pt}
    & \hspace{-4pt} -0.182 \hspace{-5pt}
    & \hspace{-4pt} -0.189 \hspace{-5pt}
    & \hspace{-4pt} -0.206 \hspace{-5pt}
    & \hspace{-4pt} -0.182 \hspace{-4pt} \\
    Issue-adjusted IPM (with labeled LDA)
    & \hspace{-4pt} \textbf{-0.208} \hspace{-5pt}
    & \hspace{-4pt} \textbf{-0.209} \hspace{-5pt}
    & \hspace{-4pt} \textbf{-0.181} \hspace{-5pt}
    & \hspace{-4pt} \textbf{-0.188} \hspace{-5pt}
    & \hspace{-4pt} \textbf{-0.205} \hspace{-5pt}
    & \hspace{-4pt} \textbf{-0.180} \hspace{-4pt} \\
    Issue-adjusted IPM (with direct labels)
    & \hspace{-4pt} \textbf{-0.208} \hspace{-5pt}
    & \hspace{-4pt} \textbf{-0.209} \hspace{-5pt}
    & \hspace{-4pt} -0.182 \hspace{-5pt}
    & \hspace{-4pt} -0.189 \hspace{-5pt}
    & \hspace{-4pt} -0.206 \hspace{-5pt}
    & \hspace{-4pt} -0.181 \hspace{-4pt} \\
    Standard LDA
    & \hspace{-4pt} \textbf{-0.208} \hspace{-5pt}
    & \hspace{-4pt} -0.210 \hspace{-5pt}
    & \hspace{-4pt} -0.184 \hspace{-5pt}
    & \hspace{-4pt} -0.189 \hspace{-5pt}
    & \hspace{-4pt} -0.207 \hspace{-5pt}
    & \hspace{-4pt} -0.181 \hspace{-4pt} \\
    Issue-adjusted IPM (with permuted issues)
    & \hspace{-4pt} -0.210 \hspace{-5pt}
    & \hspace{-4pt} -0.210 \hspace{-5pt}
    & \hspace{-4pt} -0.183 \hspace{-5pt}
    & \hspace{-4pt} -0.203 \hspace{-5pt}
    & \hspace{-4pt} -0.211 \hspace{-5pt}
    & \hspace{-4pt} -0.186 \hspace{-4pt} \\
    \hline
    \end{tabular}
    \center
    \textbf{Heldout log likelihood of House votes} \\
    \begin{tabular}{ccccccc}
    \hline
    \hline
    Traditional ideal point model (IPM) & \hspace{-4pt} -0.168 \hspace{-5pt}
    & \hspace{-4pt} -0.154 \hspace{-5pt}
    & \hspace{-4pt} -0.096 \hspace{-5pt}
    & \hspace{-4pt} -0.120 \hspace{-5pt}
    & \hspace{-4pt} -0.090 \hspace{-5pt}
    & \hspace{-4pt} -0.077 \hspace{-4pt} \\
    Issue-adjusted IPM (with labeled LDA)
    & \hspace{-4pt} \textbf{-0.167} \hspace{-5pt}
    & \hspace{-4pt} \textbf{-0.151} \hspace{-5pt}
    & \hspace{-4pt} -0.095 \hspace{-5pt}
    & \hspace{-4pt} -0.118 \hspace{-5pt}
    & \hspace{-4pt} -0.089 \hspace{-5pt}
    & \hspace{-4pt} -0.076 \hspace{-4pt} \\
    Issue-adjusted IPM (with direct labels)
    & \hspace{-4pt} \textbf{-0.167} \hspace{-5pt}
    & \hspace{-4pt} \textbf{-0.151} \hspace{-5pt}
    & \hspace{-4pt} \textbf{-0.094} \hspace{-5pt}
    & \hspace{-4pt} \textbf{-0.117} \hspace{-5pt}
    & \hspace{-4pt} \textbf{-0.088} \hspace{-5pt}
    & \hspace{-4pt} \textbf{-0.075} \hspace{-4pt} \\
    Standard LDA
    & \hspace{-4pt} -0.171 \hspace{-5pt}
    & \hspace{-4pt} -0.154 \hspace{-5pt}
    & \hspace{-4pt} -0.097 \hspace{-5pt}
    & \hspace{-4pt} -0.121 \hspace{-5pt}
    & \hspace{-4pt} -0.091 \hspace{-5pt}
    & \hspace{-4pt} -0.078 \hspace{-4pt} \\
    Issue-adjusted IPM (with permuted issues)
    & \hspace{-4pt} -0.167 \hspace{-5pt}
    & \hspace{-4pt} -0.155 \hspace{-5pt}
    & \hspace{-4pt} -0.096 \hspace{-5pt}
    & \hspace{-4pt} -0.122 \hspace{-5pt}
    & \hspace{-4pt} -0.090 \hspace{-5pt}
    & \hspace{-4pt} -0.077 \hspace{-4pt} \\
    \hline
  \end{tabular}
  \normalsize
  \label{table:session_comparison}
\end{figure}

We summarize the results in Table~\ref{table:session_comparison}. In
all chambers in both Congresses, the issue-adjusted model represents
heldout votes with higher log-likelihood than an ideal point model.
Further, every permutation represented votes with lower log-likelihood
than the issue-adjusted model.  In most cases they were also lower
than an ideal point model.  These tables validate the additional
complexity of the issue-adjusted ideal point model.







\section{Exploring Issues and Lawmakers}
\label{section:lawmakers}
\label{section:issues}

In the previous section, we demonstrated that the issue-adjusted IPM
gives a better fit to roll call data than the traditional ideal point
model.  While we used prediction to validate the model, we emphasize
that it is primarily an exploratory tool.  As for the traditional
ideal point model, it is useful for summarizing and characterizing
roll call data.  In this section, we demonstrate how to use the
approximate posterior to explore a collection of bills, lawmakers, and
votes through the lens of the issue-adjusted model.

We will focus on the 111th Congress (2009-2010).  First, we show on
which issues the issue-adjusted model best fits.  We then discuss
several specific lawmakers, showing voting patterns that identify
lawmakers who transcend their party lines.  We finally describe
procedural cartel theory~\citep{cox:2002}, which explains why certain
lawmakers have such different preferences on procedural issues like
\emph{congressional sessions} than substantive issues like as
\emph{finance}.





\subsection{Issues Improved by Issue Adjustment}
\label{section:issue_improvements}

Which issues give the issue-adjusted model an edge over the
traditional model?  We measured this with a metric we will refer to as
\emph{issue improvement}.  Issue improvement is the weighted
improvement in log likelihood for the issue-adjusted model relative to
the traditional model.  We formalize this by defining the log
likelihood of each lawmaker's vote
\begin{equation}
  J_{ud} = 1_{\{v_{ud} = \mbox{yes}\}} p - \log(1 + \exp(p)),
\end{equation}
where $p = (x_u + \bm z_{u}^T \bm \theta_d ) a_d + b_d$ is the
log-odds of a vote under the issue-adjusted voting model.  We also
measure the corresponding log-likelihood $I_{ud}$ under the ideal
point model, using $p=x_u a_d + b_d$. The improvement of issue $k$ is
then the sum of the improvement in log-likelihood, weighted by how
much each vote represents issue $k$:
\begin{equation}
  \label{equation:likelihood_improvement}
  \mbox{Imp}_k = \frac{\sum_{v_{ud}} \bm \theta_{d_v k} (J_{ud} - I_{ud}) }
       { \sum_{v_{ud}} \bm \theta_{d_v k} }.
     \end{equation}
A high value of $\mbox{Imp}_k$ indicates that issue $k$ is associated
with an increase in log-likelihood, while a low value is associated
with a decrease in log-likelihood.

We measured this for each issue in the 111th House.  As this was an
empirical question about the entire House, we fit the model to all
votes (in contrast to the analysis above, which fit the model to five
out of six folds, for each of the six folds).

\begin{figure}
  \center
  \includegraphics[width=0.9\textwidth]{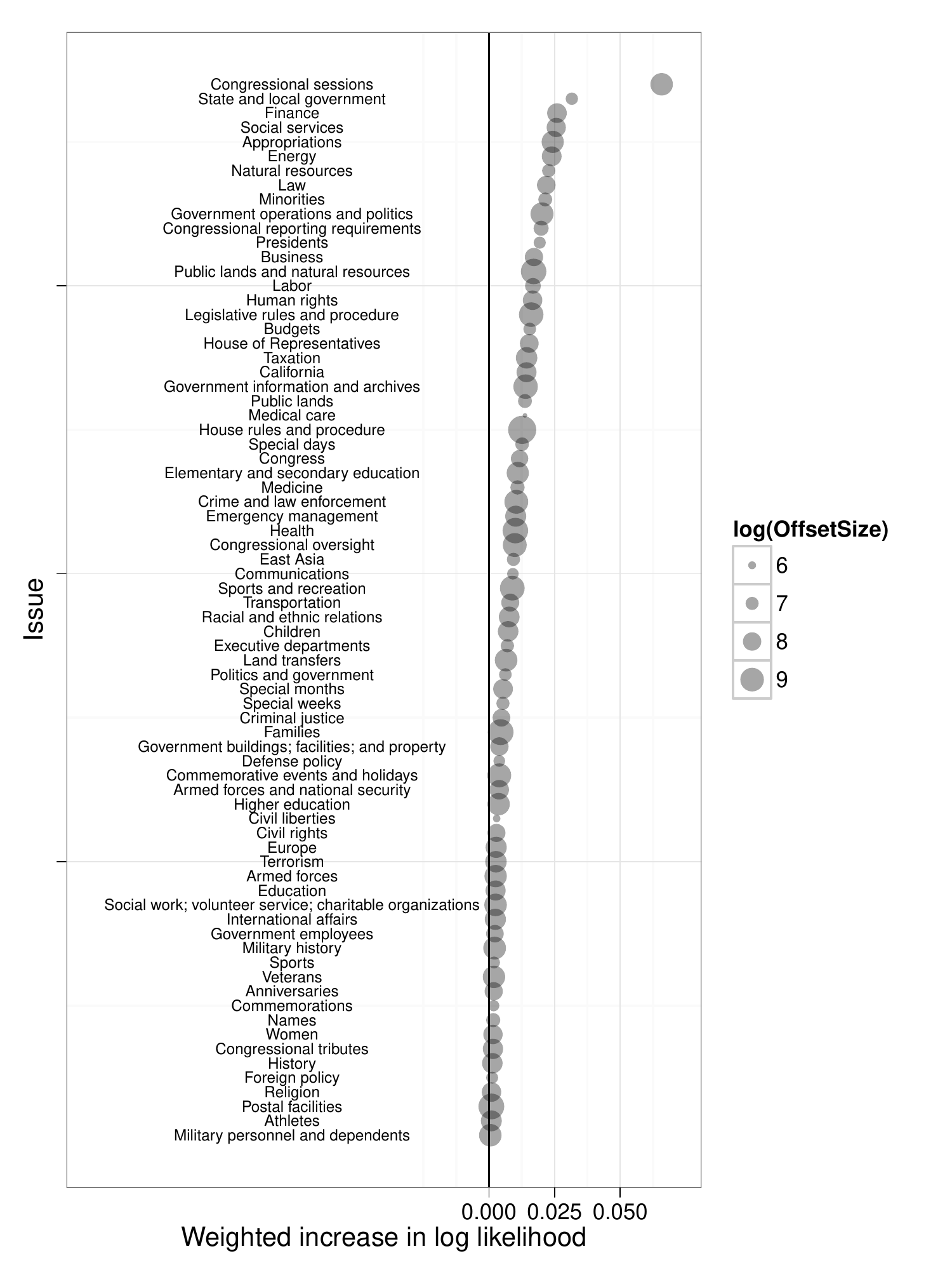}
  \caption{Log-likelihood increases when using adjusted ideal points
    most for procedural and strategic votes and less for issues
    frequently discussed during elections.  $\mbox{Imp}_k$ is shown on
    the x-axis, while issues are spread on the y-axis for display.
    The size of each issue $k$ is proportional to the logarithm of the
    weighted sum $\sum_{v_{ud}} \bm \theta_{dk}$ of votes about the
    issue.}
  \label{figure:issue_improvements}
\end{figure}
We illustrate $\mbox{Imp}_k$ for a all issues in
Figure~\ref{figure:issue_improvements}.  All issues increased
log-likelihood; those associated with the greatest increase tended to
be related to procedural votes.  For example, \emph{women},
\emph{religion}, and \emph{military personnel} issues are nearly
unaffected by lawmakers' offsets.  For those issues, a global
political spectrum (i.e., a single dimension) capably explains the
lawmakers' positions.

\subsection{Exploring the Issue Adjustments}
\label{section:party_bias}

The purpose of our model is to adjust lawmaker's ideal points
according to the issues under discussion.  In this section, we
demonstrate a number of ways to explore this information.

\begin{figure}[t]
  \center
  \includegraphics[width=0.82\textwidth]{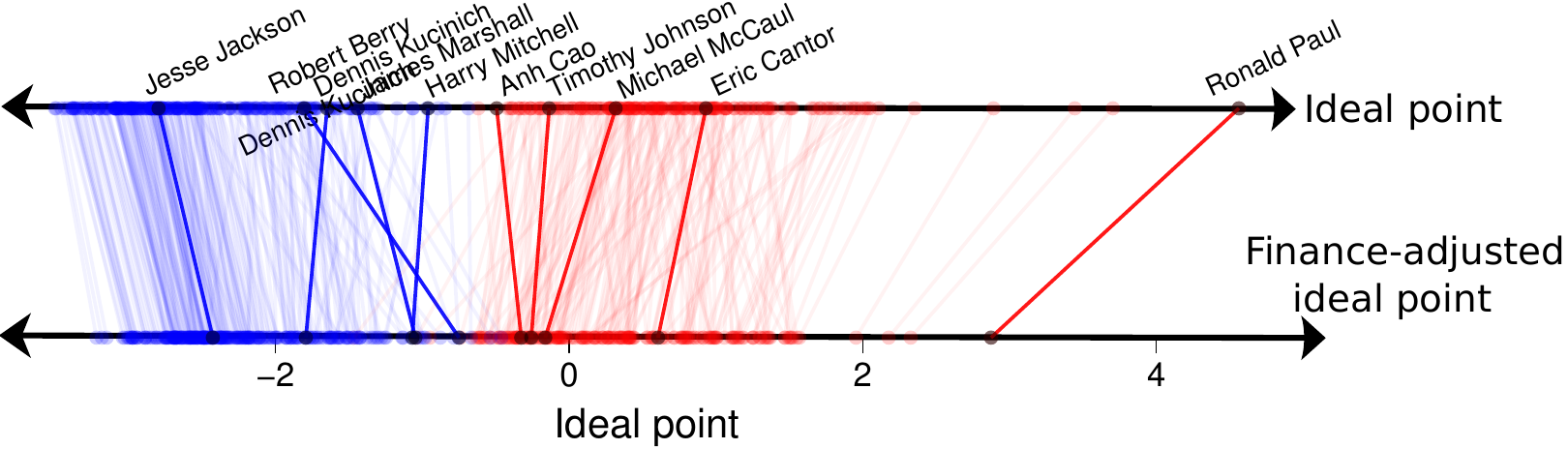}
  \includegraphics[width=0.82\textwidth]{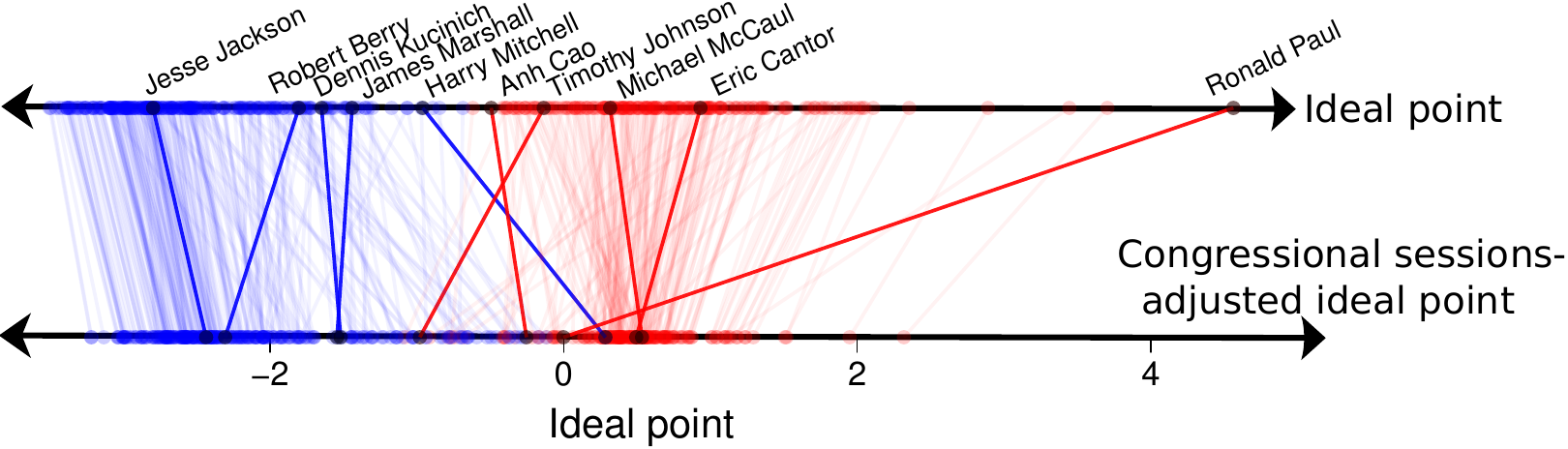}
  \caption{Ideal points $x_u$ and issue-adjusted ideal points $x_u +
    z_{uk}$ from the 111th House for the substantive issue
    \emph{finance} and the procedural issue \emph{congressional
      sessions}. Democrats are blue and Republicans are red.  Votes
    about \emph{finance} and \emph{congressional sessions} were better
    fit using issue-adjusted ideal points.  For procedural votes such
    as \emph{congressional sessions}, lawmakers become more
    polarized by political party, behavior predicted by procedural
    cartel theory \citep{cox:1993}. }
  \label{figure:issue_improvements_ideals}
\end{figure}
\begin{figure}
  \center
  \includegraphics[width=1.0\textwidth]{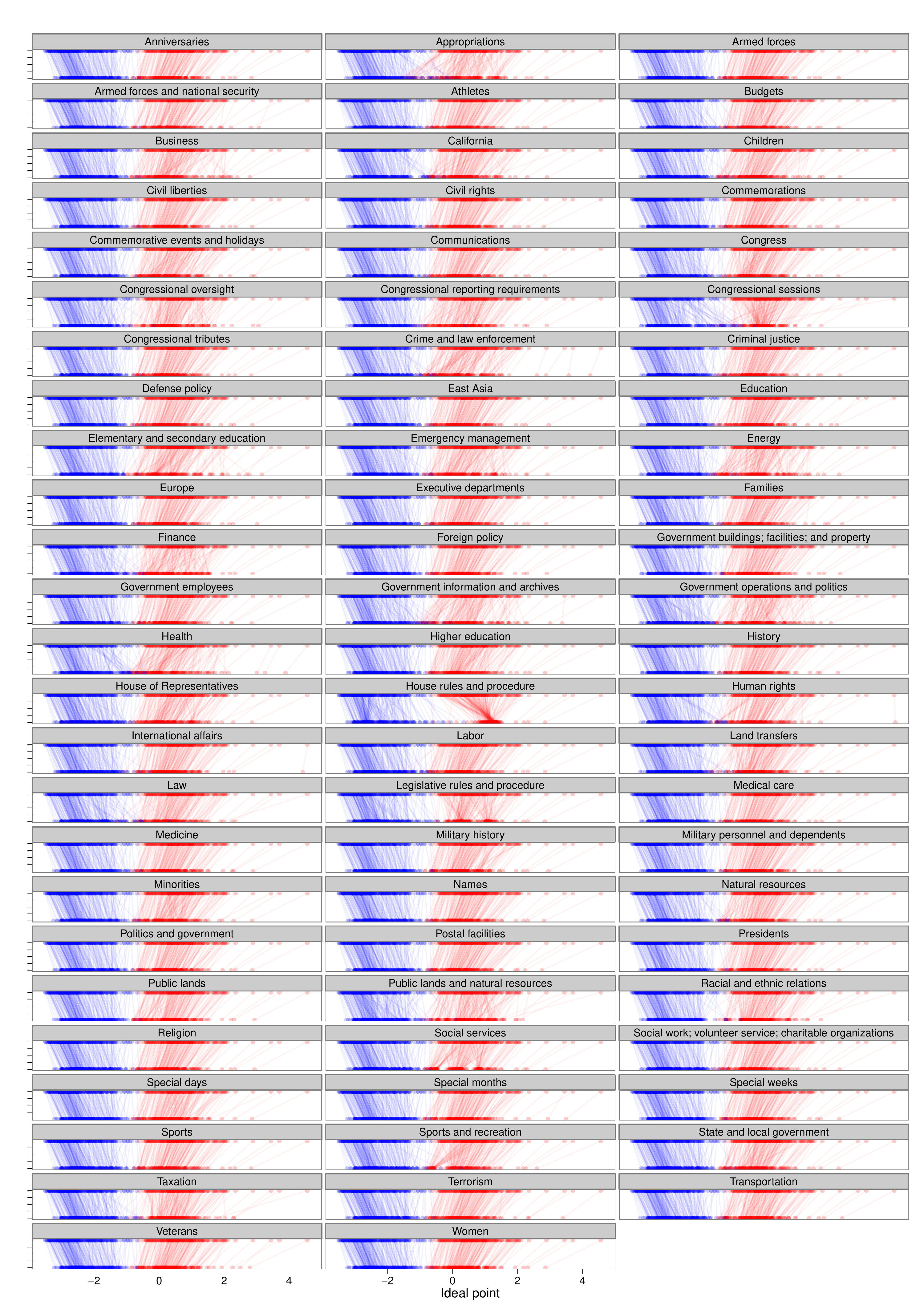}
  \caption{Ideal points $x_u$ and issue-adjusted ideal points $x_u +
    z_{uk}$ from the 111th House for all issues. }
  \label{figure:all_issue_adjusted_ideals}
\end{figure}

We begin with a brief summary of the main information obtained by this
model. During posterior inference, we jointly estimate the mean
$\tilde x_u, \tilde z_u$ of all lawmakers' positions lawmakers'
issue-adjusted ideal points.  These issue adjustments $\tilde z_u$
adjust how we expect lawmakers to vote given their un-adjusted ideal
points $\tilde x$. We illustrate this for \emph{finance} (a
substantive issue) and \emph{congressional sessions} (a procedural
issue) in \myfig{issue_improvements_ideals} and summarize all issues
in \myfig{all_issue_adjusted_ideals}.  Upon a cursory inspection, it
is clear that in some issues, lawmakers' adjustments are relatively
sparse, i.e. only a few lawmakers' adjustments are interesting.  In
other issues---such as the procedural issue \emph{congressional
 sessions}---these adjustments are more systemic.

\begin{figure}
  \center
  \begin{tabular}{cc}
    \Large \textbf{Democrat} & \Large \textbf{Republican} \normalsize \\
    \includegraphics[width=0.45\textwidth]{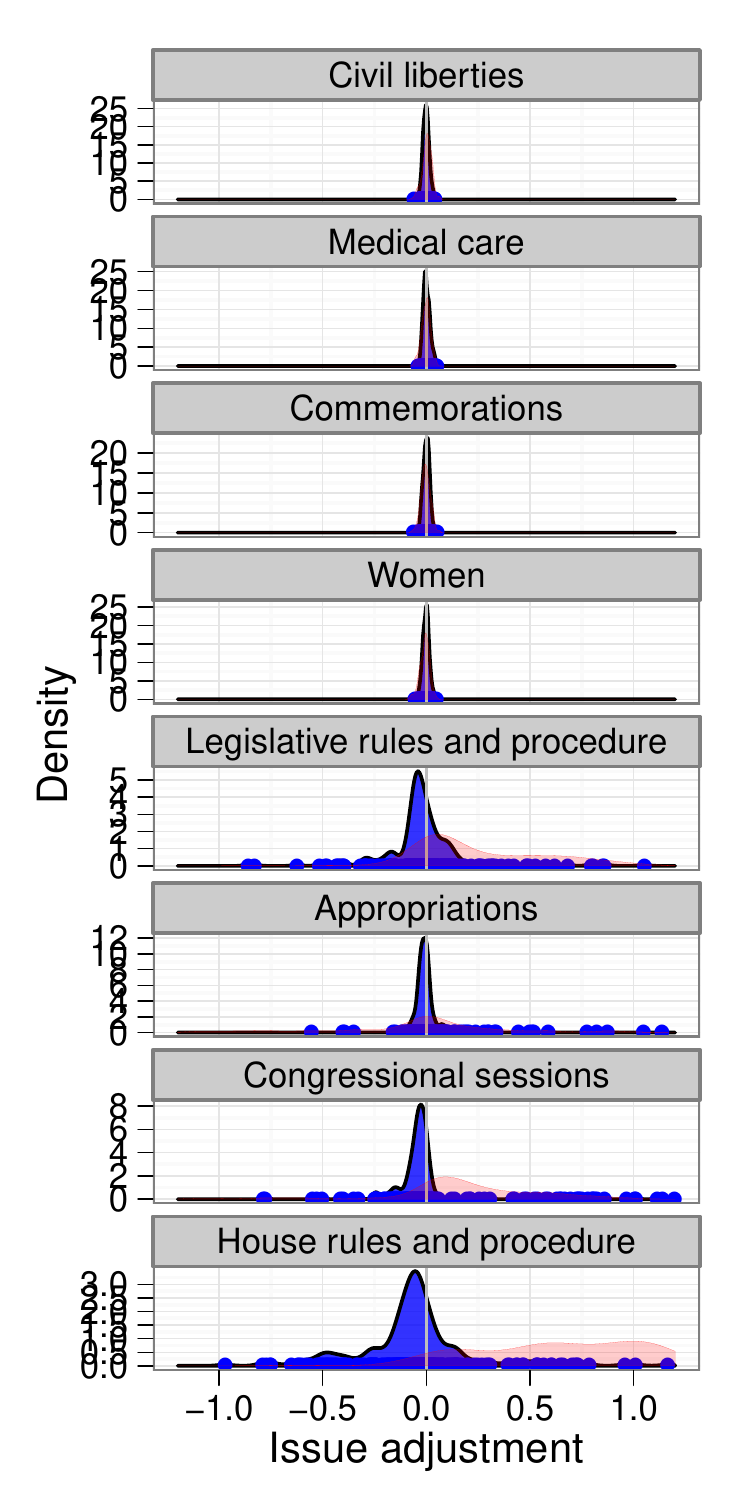} &
    \includegraphics[width=0.45\textwidth]{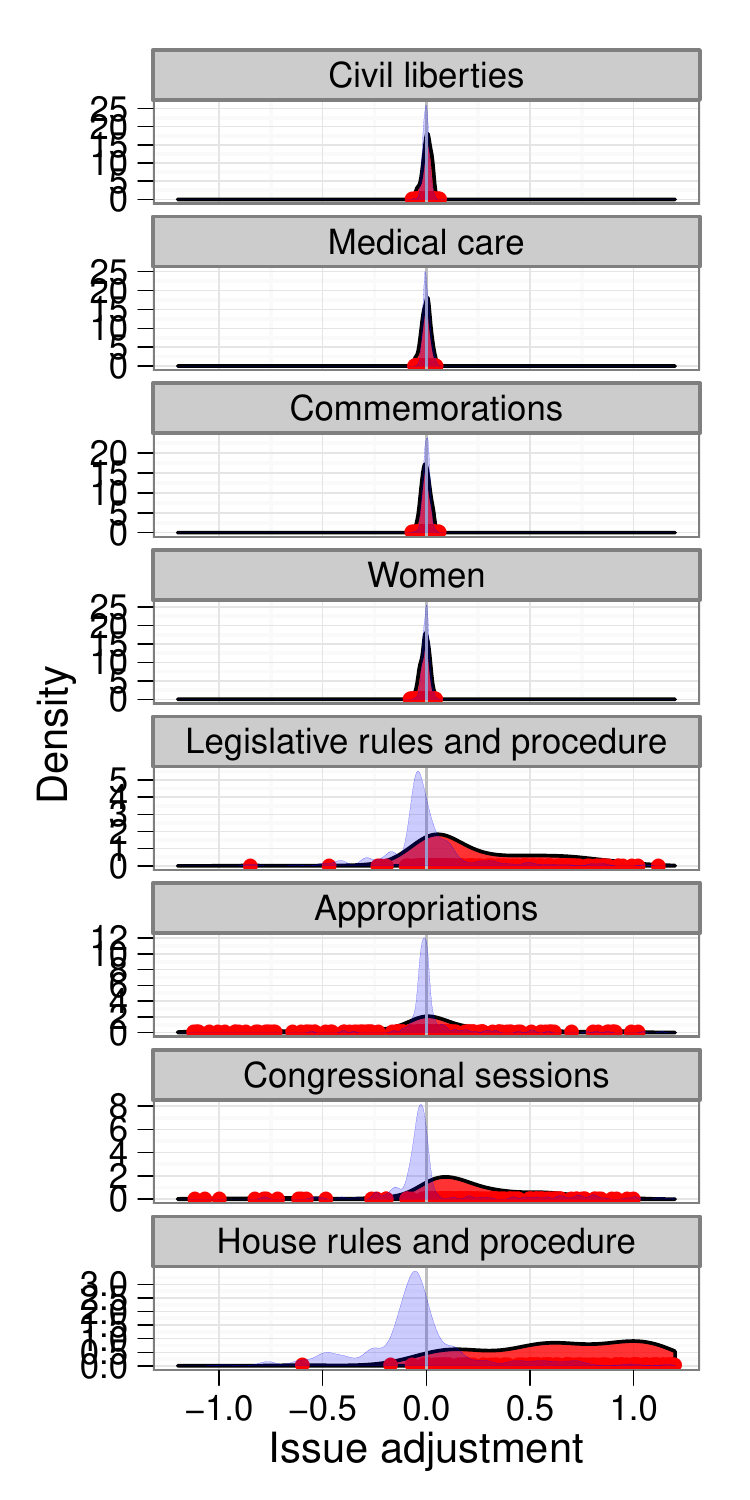} \\
  \end{tabular}
  \caption{Histogram of issue adjustments for selected
    issues. Democrats are in the left column, and Republicans are in
    the right column. Both Democrats and Republicans tend to have
    small issue adjustments for traditional issues.  Their issue
    adjustments differ substantially for procedural issues. A
    more-dispersed distribution of issue adjustments does not mean
    that these lawmakers tend to feel differently from one another
    about these issues.  Instead, it means that lawmakers deviate from
    their ideal points more.  }
  \label{figure:issue_adjustment_distribution}
\end{figure}

\paragraph{Adjustments by issue and party.}
\myfig{issue_adjustment_distribution} illustrates the distribution
across lawmakers of the posterior issue adjustments (denoted $\tilde
z_{uk}$) for issues with the highest and lowest variance.
This figure shows the distribution for the four issues with the
greatest variation in $\tilde z_{uk}$ (across lawmakers) and the four
issues with the least variation.  Note the systematic bias in
Democrats' and Republicans' issue preferences: they become more
partisan on certain issues, particularly procedural ones.





\paragraph{Controlling for ideal points.}
\label{section:conditional_offsets}
We found that posterior issue adjustments can correlate with the ideal
point of the lawmaker---for example, a typical Republican tends to
have a Republican offset on taxation. In some settings, we are more
interested in understanding when a Republican deviates from
behavior suggested by her ideal point.  We can shed light on this
systemic issue bias by explicitly controlling for it.  To do this, we
fit a regression for each issue $k$ to explain away the effect of a
lawmaker's ideal point $\bm x_u$ on her offset $\bm z_{uk}$:
\[
  \bm z_{k} = \beta_k \bm X + \bm \varepsilon,
\]
where $\beta_k \in \mathbb{R}$.  Instead of evaluating a lawmaker's
observed offsets, we use her residual $\hat z_{uk} = \bm z_{uk} -
\beta_k \bm x_u$, which we call the \textit{corrected issue
  adjustment}.  By doing this, we can evaluate lawmakers in the
context of other lawmakers who share the same ideal points: a positive
offset $\hat z_{uk}$ for a Democrat means she tends to vote more
conservatively about issue $k$ than others with the same ideal point
(most of whom are Democrats).



Most issues had only a moderate relationship to ideal points.
\emph{House rules and procedure} was the most-correlated with ideal
points, moving the adjusted ideal point $\beta_k=0.26$ right for every
unit increase in ideal point. \emph{public land and natural resources}
and \emph{taxation} followed at a distance, moving an ideal point
$0.04$ and $0.025$ respectively with each unit increase in ideal
point.  \emph{health}, on the other hand, moved lawmakers
$\beta_k=0.04$ left for every unit increase in ideal point.  At the
other end of the spectrum, the issues \emph{women}, \emph{religion},
and \emph{military personnel} were nearly unaffected by lawmakers'
offsets.

\paragraph{Extreme lawmakers.}



We next use these corrected issue adjustments to identify lawmakers'
exceptional issue preferences. To identify adjustments which are
significant, we turn again to the same nonparametric check described
in the last section: permute issue vectors' document labels,
i.e. $(\bm \theta_1, \ldots, \bm \theta_D) \mapsto (\bm
\theta_{\pi_i(1)} \ldots \bm \theta_{\pi_i(D)})$, and refit lawmakers'
adjustments using both the original issue vectors and permuted issue
vectors, for permutations $\pi_1, \ldots, \pi_{20}$.  We then compare
a corrected issue adjustment $\hat z_{uk}$'s absolute value with
corrected issue adjustments estimated with permuted issue vectors $\bm
\theta_{\pi_i(d) k}$.  This provides a nonparametric method for
finding issue adjustments which are more extreme than expected by
chance: an extreme issue adjustment has a greater absolute value than
all of its permuted counterparts.  We use these to discuss several
unique lawmakers.

\begin{figure}
  \center
  \begin{tabular}{cc}
    \includegraphics[width=0.5\textwidth]{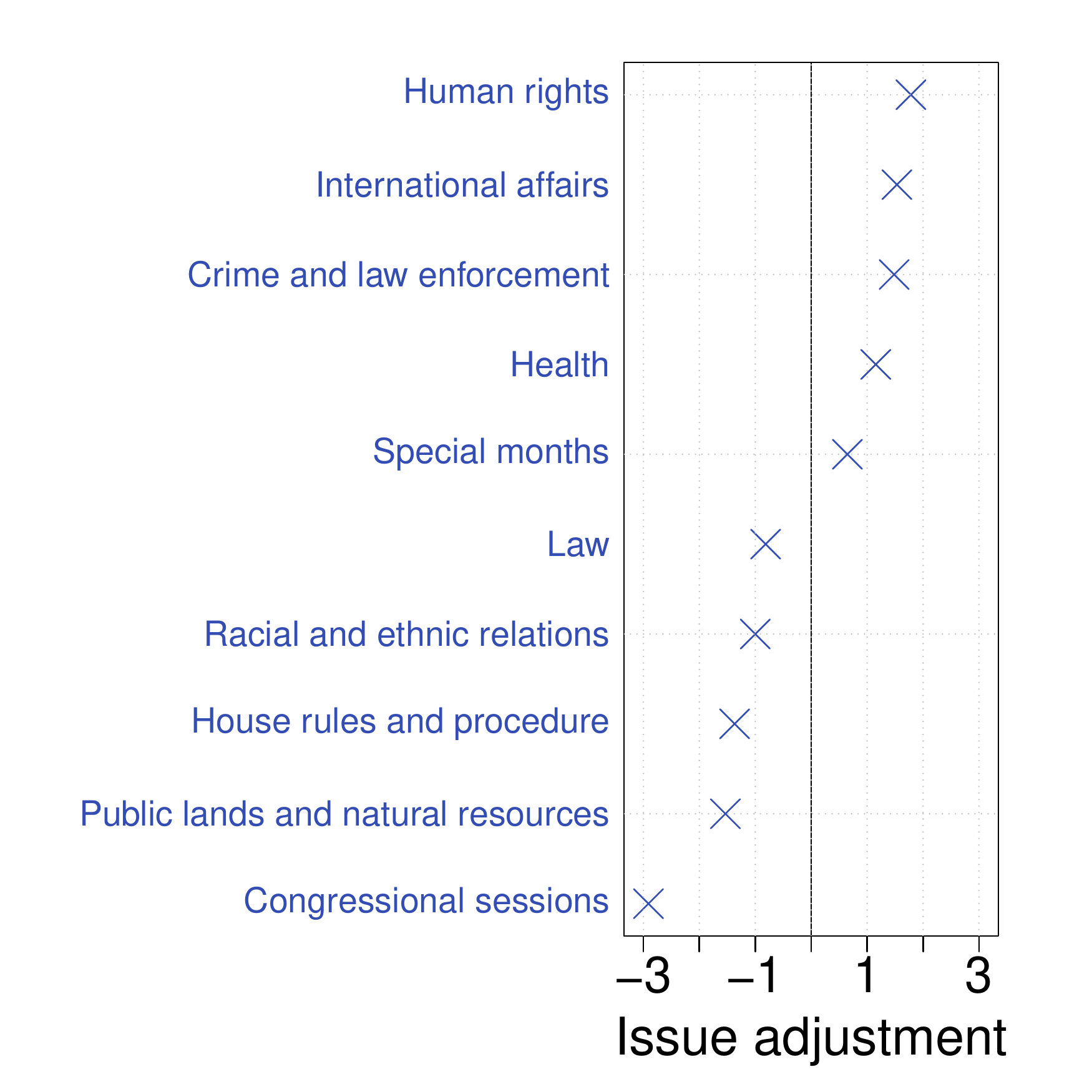} &
    \includegraphics[width=0.5\textwidth]{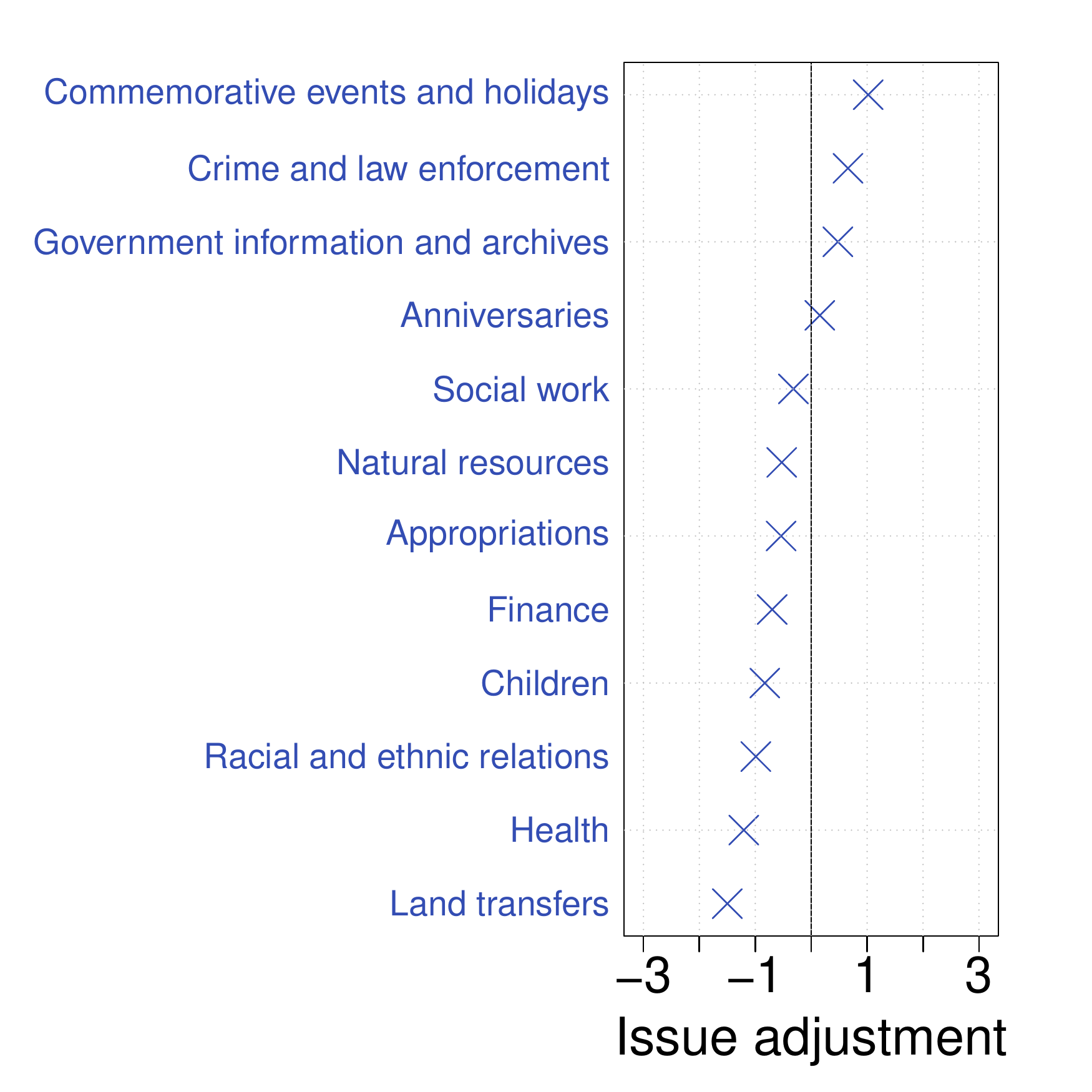} \\
    Ronald Paul (House Republican) & Donald Young (House Republican) \\
  \end{tabular}
    \caption{Significant issue adjustments for exceptional senators in
  Congress 111.  Each illustrated issue is significant to $p <
  0.05$ by a permutation test.}
  \label{figure:significant_offsets}
\end{figure}

Using corrected issue adjustments, we identified several of the
most-unique lawmakers.  We focused this analysis on votes from
2009-2010, the most recent full session of Congress, using
$\lambda=1$.  We fit the variational approximation to all votes in the
House and computed lawmakers' corrected issue adjustments $\hat
z_{uk}$, which are conditioned on their ideal points as described in
\mysec{conditional_offsets}.
Figure~\ref{figure:significant_offsets} illustrates those issue
preferences which were significant under 20 permutation replications
($p < 0.05$) for several lawmakers from this Congress.

\begin{itemize}
  \item \textbf{Ron Paul.}
We return to Ron Paul, one of the most unique House Republicans, and a
lawmaker who first motivated this analysis.  Paul's offsets were very
extreme; he tended to vote more conservatively than expected on
\emph{health}, \emph{human rights} and \emph{international affairs}.
He voted more liberally on social issues such as \emph{racial and
  ethnic relations}, and broke with behavior expected under a
procedural cartel (congressional sessions).
The issue-adjusted training accuracy of Paul's votes increased from
83.8\% to 87.9\% with issue offsets, placing him among the two
most-improved lawmakers with this model. %

The issue-adjusted improvement $\mbox{Imp}_K$
(Equation~\ref{equation:likelihood_improvement}) when restricted to
Paul's votes indicate significant improvement in \emph{international
  affairs} and \emph{East Asia} (he tends votes against
U.S. involvement in foreign countries); \emph{congressional sessions};
\emph{human rights}; and \emph{special months} (he tends to vote
against recognition of months as special holidays).

\item \textbf{Donald Young.}  One of the most exceptional legislators
  in the 111th House was Donald Young, Alaska Republican.  Young stood
  out most in a topic used frequently in House bills about naming
  local landmarks. In many cases, Young voted against the majority of
  his party (and the House in general) on a series of largely symbolic
  bills and resolutions.  For example, in the \emph{commemorative
    events and holidays} topic, Young voted (with only two other
  Republicans and against the majority of the House) \emph{not to}
  commend ``the members of the Agri-business Development Teams of the
  National Guard and the National Guard Bureau for their efforts... to
  modernize agriculture practices and increase food production in
  war-torn countries.'' %

  Young's divergent symbolic voting was also evident in a series of
  votes against naming various landmarks---such as post offices---in
  a topic about such symbolic votes. Yet Donald Young's ideal point is
  -0.35, which is not particularly distinctive (see
  Figure~\ref{figure:classic_ideal_points}): using the ideal point
  alone, we would not recognize his unique voting behavior.
\end{itemize}

\paragraph{Procedural Cartels.}
\label{section:procedural_cartel_theory}
Above we briefly noted that Democrats and Republicans become
\emph{more partisan} on procedural issues.  Lawmakers' more partisan
voting on procedural issues can be explained by theories about
partisan strategy in the House.  In this section we summarize a theory
underlying this behavior and note several ways in which it is
supported by issue adjustments.

The sharp contrast in voting patterns between procedural votes and
substantive votes has been noted and studied over the past century
\citep{fenno:1965,jones:1964,cox:1993,cox:2002}.
\citet{cox:1993} provide a summary of this behavior: ``parties in the
House---especially the majority party---are a species of
'legislative cartel' [ which usurp the power ] to make rules governing
the structure and process of legislation.''  A defining assumption
made by \cite{cox:2005} is that the majority party delegates an
agenda-setting monopoly to senior partners in the party, who set the
procedural agenda in the House.  As a result, the cartel ensures that
senior members hold agenda-setting seats (such as committee chairs)
while rank-and-file members of the party support agenda-setting
decisions.



This \emph{procedural cartel theory} has withstood tests in which
metrics of polarity were found to be \emph{greater} on procedural
votes than substantive votes~\citep{cox:1993,cox:2002,cox:2005}. We
note that issue adjustments support this theory in several ways.
First, lawmakers' systematic bias for procedural issues was
illustrated and discussed in \mysec{party_bias} (see
\myfig{issue_adjustment_distribution}): Democrats systematically lean
left on procedural issues, while Republicans systematically lean
right.  Importantly, this discrepancy is more pronounced among
procedural issues than substantive ones.  Second, lawmakers' positions
on procedural issues are more partisan than expected under the
underlying un-adjusted ideal points (see
\mysec{conditional_offsets} and
\myfig{issue_improvements_ideals}).  Finally, more extreme polarity
and improved prediction on procedural votes (see \mysec{performance}
and \myfig{bills_parameter_changes}) indicate that that issue
adjustments for procedural votes are associated with \emph{more
  extreme} party affiliation---also observed by \citet{cox:2002}.


\section{Summary}

We developed and studied the issue-adjusted ideal point model, a model
designed to tease apart lawmakers' preferences from their general
political position.  This is a model of roll-call data that captures
how lawmakers vary, issue by issue.  It gives a new way to explore
legislative data.  On a large data set of legislative history, we
demonstrated that it is able to represent votes better than a classic
ideal point model and illustrated its use as an exploratory
tool.

This work could be extended in several ways.  One of the most natural
way is to incorporate lawmakers' stated positions on issues -- which
may differ from how they actually vote on these issues; in preliminary
analyses, we have found little correlation to external sources.  We
might also study lawmakers' activities outside of voting to understand
their issue positions.  For example, lawmakers' fund-raising by
industry area might (or might not) be useful in predicting their
positions on different issues.  Additional work includes modeling how
lawmakers' positions on issues change over time, by incorporating
time-series assumptions as in \cite{martin:2002}.





\bibliography{bib}

\newpage
\setcounter{figure}{0} \renewcommand{\thefigure}{\thesection.\arabic{figure}}

\titleformat{\section}
  {\normalfont\large\uppercase}{\large APPENDIX \thesection.}{.5em}{}

\appendix

\section{Posterior inference}

In this appendix we provide additional details for \emph{A Textual Issue Model for Legislative Roll Calls}.  We begin by detailing the inference algorithm summarized
in \mysec{inference}.

\subsection{Optimizing the variational objective}

Variational bounds are typically optimized by gradient ascent or block
coordinate ascent, iterating through the variational parameters and
updating them until the relative increase in the lower bound is below
a specified threshold.  Traditionally this would require symbolic
expansion of the ELBO $\mathcal{L}_\eta = \expectq{p(x, v, \bm z, \bm
  \theta, a, b) - q_{\eta}(x, v, \bm z, a, b)}$, so that the bound can
be optimized with respect to the variational parameters $\eta$. This
expectation cannot be analytically expanded with our model.  One
solution would be to approximate this bound.  Especially when there
are many variables, however, this approximation and the resulting
optimization algorithm are complicated and prone to bugs.

Instead of expanding this bound symbolically, we update each parameter
with stochastic optimization.  We repeat these updates for each
parameter until the parameters have converged.  Upon convergence, we
use the variational means $\tilde x$ and $\bm \tilde z$ to inspect
lawmakers' issues and bill parameters $\tilde a$ and $\tilde b$ to
inspect items of legislation.

Without loss of generality, we describe how to perform the $m$th
update on the variational parameter $\tilde x$, assuming that we have
the most-recent estimates of the variational parameters $\tilde
x_{n-1}, \tilde z_{n-1}, \tilde a_{n-1},$ and $\tilde b_{n-1}$.  To
motivate the inference algorithm, we first approximate the ELBO
$\mathcal{L}$ with a Taylor approximation, which we optimize.  At the
optimum, the Taylor approximation is equal to the ELBO.

Writing the variational objective as $\mathcal{L}(\tilde x) =
\mbox{KL}(q_{\tilde x} || p)$ for notational convenience (where all
parameters in $\eta$ except $\tilde x$ are held fixed), we estimate
the KL divergence as a function of $\tilde x$ around our last estimate
$\tilde x_{m-1}$ with its Taylor approximation
\begin{align}
  \label{equation:taylor_approximation}
  \mathcal{L}(\tilde x) \approx & \mathcal{L}(\tilde x_{n-1})
  + \left( \partl{\mathcal{L}}{\tilde x} \Bigr|_{\tilde x_{n-1}} \right)^T \Delta \tilde x + \frac{1}{2} \Delta \tilde x^T \left( \partlll{\mathcal{L}}{\tilde x}{\tilde x} \Bigr|_{\tilde x_{n-1}} \right) \Delta \tilde x, \\ \nonumber
\end{align}
where $\Delta \tilde x = \tilde x - \tilde x_{n-1}$.  Once we have estimated the Taylor
coefficients (as described in the next paragraph), we can perform the update
\begin{align}
  \label{equation:taylor_update}
  \tilde x_n \gets \tilde x_{n-1} - \left( \partlll{\mathcal{L}}{\tilde x}{\tilde x} \Bigr|_{\tilde x_{n-1}} \right)^{-1} \left( \partl{\mathcal{L}}{\tilde x} \Bigr|_{\tilde x_{n-1}} \right).
\end{align}



\label{appendix:approximate_hessian}
\label{taylor_update}
\label{appendix:taylor_update}
We approximated the Taylor coefficients with Monte Carlo sampling.
Without loss of generality, we will illustrate this approximation with
the variational parameter $\tilde x$.  Let $\tilde x_{n-1}$ be the current
estimates of the variational mean, $q_{\tilde x_{n-1}}(x, \bm z, a, b)$ be
the variational posterior at this mean, and $\mathcal{L}_{\tilde x_{n-1}}$
be the ELBO at this mean.  We then approximate the gradient with Monte
Carlo samples as
\begin{align}
  \partl{\mathcal{L}_{\tilde x}}{\tilde x} \Bigr|_{\tilde x_{n-1}} & = \partl{}{\tilde x} \int q_{\tilde x}(x, \bm z, a, b) ( \log p(x, \bm z, a, b, v) - \log q_{\tilde x}(x, \bm z, a, b) ) dx d\bm z da db \Bigr|_{\tilde x_{n-1}} \\ \nonumber
  & = \int \partl{}{\tilde x} \left( q_{\tilde x}(x) ( \log p(x, \bm z, a, b, v) - \log q_{\tilde x}(x, \bm z, a, b) ) \right) d \tilde x \Bigr|_{\tilde x_{n-1}} \\ \nonumber
  & = \int q_{\tilde x}(x) \partl{\log q_{\tilde{x}(x)}}{\tilde x} \Bigr|_{\tilde x_{n-1}} ( \log p(x, \bm z, a, b, v) - \log q_{\tilde x}(x, \bm z, a, b) ) d \tilde x \\ \nonumber
  & \approx \displaystyle \frac{1}{M} \sum_{m=1}^M \Bigg( \left( \partl{\log q_{\tilde x}(x_{n-1,m}, z_{n-1,m}, a_{n-1,m}, b_{n-1,m})}{\tilde x} \Bigr|_{ \tilde x_{n-1} } \right) \\ \nonumber
  & \hspace{75pt} \times \big( \log p(x_{n-1,m}, z_{n-1,m}, a_{n-1,m}, b_{n-1,m}, v) \\ \nonumber
  & \hspace{95pt} - C - \log q_{\tilde x_{n-1}}(x_{n-1,m}, z_{n-1,m}, a_{n-1,m}, b_{n-1,m}) \big) \Bigg), \nonumber
\end{align}
where we have taken the gradient through the integral using Leibniz's
rule and used $M$ samples from the current estimate of the variational
posterior.  The second Taylor coefficient is straightforward to derive
with similar algebra:
\begin{align}
  \partlll{\mathcal{L_{\tilde x}}}{\tilde x}{\tilde x} \Bigr|_{\tilde x_{n-1}}
  \approx \displaystyle \frac{1}{M} \sum_{m=1}^M
  \Bigg( & \hspace{0pt} \left( \partl{\log q_{n-1}(x_{n-1,m})}{\tilde x} \Bigr|_{\tilde x_{n-1}} \right) 
 \left( \partl{\log q_{m-1}(x_{n-1,m})}{\tilde x} \Bigr|_{\tilde x_{n-1}} \right)^T \\ \nonumber
 &  \hspace{10pt} \times \big( \log p(x_{n-1,m}, \bm z_{n-1,m}, a_{n-1,m}, a_{n-1,m}, v) \\ \nonumber
 & \hspace{30pt} - C - \log q_{\tilde x_{n-1}}(x_{n-1,m}, z_{n-1,m}, a_{n-1,m}, b_{n-1,m}) - 1 \big), \\ \nonumber
  & \hspace{0pt} + \left( \bigg( \partlll{\log q_{n-1}(x_{n-1,m})}{\tilde x}{\tilde x} \Bigr|_{\tilde x_{n-1}} \right) \\ \nonumber
 & \hspace{10pt} \times \big( \log p(x_{n-1,m}, \bm z_{n-1,m}, a_{n-1,m}, b_{n-1,m}, v) \\ \nonumber
  & \hspace{30pt} - C - \log q_{\tilde x_{n-1}}(x_{n-1,m}, z_{n-1,m}, a_{n-1,m}, b_{n-1,m}) \big) \bigg) \Bigg),
\end{align}
where we increase $M$ as the model converges.  Note that $C$ is a free
parameter that we can set without changing the final solution. We set
$C$ to the average of $\log p(x_{n-1,m} | ...) - \log
q_{n-1}(x_{n-1,m})$ across the set of $M$ samples.


\paragraph{Quasi-Monte Carlo samples.}
Instead of taking \emph{iid} samples from the variational distribution
$q_{M-1}$, we used quasi-Monte Carlo sampling
\cite{niederreiter:1992}.  By taking non-\emph{iid} samples from
$q_{m-1}$, we are able to decrease the variance around estimates of
the Taylor coefficients.  To select these samples, we took $M$
equally-spaced points from the unit interval, passed these through the
inverse CDF of the variational Gaussian $q_{n-1}(x)$, and used the
resulting values as samples. Note that these samples produce a biased
estimate of Equation~\ref{equation:taylor_approximation}.  This bias
decreases as $N \rightarrow \infty$.

When we update the variational parameter $\tilde x_u$, we do not need
to sample all random variables, but we do need a sample of all random
variables in the Markov blanket of $x_u$. The cumulative distribution
is of course ill-defined for multivariate distributions, so the method
in the last paragraph is not quite enough.  For a quasi-Monte Carlo
sample from the multivariate distribution of $x_u$'s Markov blanket,
we selected $M$ samples using the method in the previous paragraph for
each marginal in the Markov blanket of $x_u$.  We then permuted each
variable's samples and combined them for $M$ multivariate samples $\{
x_{n-1, m}, \ldots, b_{n-1, m} \}_m$ from the current estimate
$q_{n-1}$ of the variational distribution.

\paragraph{Estimating $\partl{\log q_m}{x}$.}
We estimate the gradients of $\log q$ above based on the distribution
of the variational marginals.  We have defined the variational
distribution to be factorized Gaussians, so these take the form
\begin{align}
  \label{equation:variational_marginal_gradients}
  \partl{\log q_{n-1}(x_{n-1,m})}{\tilde x} \Bigr|_{\tilde x_{n-1}} = & \frac{x_{n-1,m} - \tilde x_{n-1}}{ \sigma_x^2} \\ \nonumber
  \partll{\log q_{n-1}(x_{n-1,m})}{\tilde x} \Bigr|_{\tilde x_{n-1}} = & - \frac{1}{ \sigma_x^2 }. \\ \nonumber
\end{align}

We finally address practical details of implementing issue-adjusted
ideal points.

\subsection{Algorithmic parameters.}
We fixed the variance $\sigma_x^2$ to $\exp(-5)$.  Allowing
$\sigma_x$ to vary freely provides a better variational bound at the
expense of accuracy.  This happens because the issue-adjusting model
would sometimes fit poor means to some parameters when the posterior
variance was large: there is little penalty for this when the variance
is large.  Low posterior variance $\sigma_x^2$ is similar to a
non-sparse MaP estimate.


These updates were repeated until the exponential moving average
$\Delta_{\mbox{est},i} \gets 0.8 \Delta_{\mbox{est},{i-1}} + 0.2
\Delta_{\mbox{obs},{i}}$ of the change in KL divergence dropped below
one and the number $N$ of samples passed 500.  If the moving average
dropped below one and $N < 500$, we doubled the number of samples.

When performing the second-order updates described in
Section~\ref{section:inference}, we skipped variable updates when the
estimated Hessian was not positive definite (this disappeared when
sample sizes grew large enough).  We also limited step sizes to 0.1
(another possible reason for smaller coefficients).

\subsection{Hyperparameter settings}
\label{appendix:hyperparameters}
The most obvious parameter in the issue voting model is the
regularization term $\lambda$. 

The main parameter in the issue-adjusted model is the regularization
$\lambda$, which is shared for all issue adjustments. The Bayesian
treatment described in the Inference section of this paper
demonstrated considerable robustness to overfitting at the expense of
precision.  With $\lambda=0.001$, for example, issue adjustments
$z_{uk}$ remained on the order of single digits, while experiments
with MaP estimates yielded adjustment estimates over 100.

We report the effect of different $\lambda$ by fitting the
issue-adjusted model to the 109th Congress (1999-2000) of the House
and Senate for a range $\lambda=0.0001, \ldots, 1000$ of
regularizations.  We performed $6$-fold cross-validation, holding out
one sixth of votes in each fold, and calculated average log-likelihood
$\sum_{v_{ud} \in V_{\mbox{\tiny heldout}}} \log p(v_{ud} | \tilde
x_u, \bm \tilde z_u, \tilde a_d, \tilde b_d)$ for votes
$V_{\mbox{\tiny heldout}}$ in the heldout set. Following the algorithm
described in \mysec{inference}, we began with $M=21$ samples to
estimate the approximate gradient (\myeq{taylor_approximation}) and
scaled it by 1.2 each time the Elbo dropped below a threshold, until
it was 500. We also fixed variance $\sigma_x^2, \sigma_z^2,
\sigma_a^2, \sigma_b^2=\exp({-}5)$.  We summarize these results in
Table~\ref{table:supp_lambda_comparison}.

The variational implementation generalized well for the entire range,
representing votes best in the range $1 \le \lambda \le 10$.
Log-likelihood dropped modestly for $\lambda < 1$.  In the worst case,
log-likelihood was -0.159 in the House (this corresponds with 96\%
heldout accuracy) and -0.242 in the Senate (93\% heldout accuracy).

We recommend a modest value of $\lambda=1$, and no greater than
$\lambda=10$.  At this value, the model outperforms ideal points in
validation experiments on both the House and Senate, for a range of
Congresses.

\begin{figure}
 \center \small
  \center \textbf{109th U.S. Senate sensitivity to $\lambda$}
  \begin{tabular}{ccccccccc}
    \hline
    \hline
    Model & \multicolumn{8}{c}{Lambda} \\
    \hline
    & 1e-4
    &  1e-3 
    &  1e-2 
    &  1e-1 
    &  1 
    &  10 
    &  100
    &  1000  \\
    \hline
    Ideal
    &  -0.188 
    &  -0.189 
    &  -0.189 
    &  -0.189
    &  -0.189 
    &  -0.190 
    &  -0.189
    &  -0.189  \\ 
    Issue (LDA)
    & -0.191 
    & -0.191
    & -0.188 
    & -0.186 
    & -0.188 
    & -0.189
    & -0.189 
    & 0.198 \\ 
     Permuted Issue 
    &  -0.242 
    &  -0.245 
    &  -0.231 
    &  -0.221 
    &  -0.204 
    &  -0.208 
    &  -0.208 
    &  -0.208  \\
    \hline
  \end{tabular}
  \center \textbf{109th U.S. House sensitivity to $\lambda$}
  \begin{tabular}{ccccccccc}
    \hline
    \hline
    Model & \multicolumn{8}{c}{Lambda} \\
    \hline
    & 1e-4
    &  1e-3 
    &  1e-2 
    &  1e-1 
    &  1 
    &  10 
    &  100
    &  1000  \\
    \hline
    Ideal
    &  -0.119 
    &  -0.119 
    &  -0.119 
    &  -0.119
    &  -0.120 
    &  -0.119 
    &  -0.119
    &  -0.119  \\ 
    Issue (LDA)
    & -0.159 
    & -0.159
    & -0.158 
    & -0.139 
    & -0.118 
    & -0.119
    & -0.119 
    & 0.119 \\ 
     Permuted Issue 
    &  -0.191 
    &  -0.192 
    &  -0.189 
    &  -0.161 
    &  -0.122 
    &  -0.120 
    &  -0.120 
    &  -0.120  \\
    \hline
  \end{tabular}
  \normalsize
  \caption{Average log-likelihood of heldout votes by regularization
    $\lambda$. Log-likelihood was averaged across folds using six-fold
    cross validation for Congress 109 (2005-2006).  The variational
    distribution represented votes with higher heldout log-likelihood
    than traditional ideal points for $1 \le \lambda \le 10$. In a
    model fit with permuted issue labels (Perm. Issue), heldout
    likelihood of votes was worse than traditional ideal points for
    all regularizations $\lambda$.
  }
  \label{table:supp_lambda_comparison}
\end{figure}



\section{Corpus Preparation}
\subsection{Issue labels}
\label{appendix:topics}

In the empirical analysis, we used issue labels obtained from the
Congressional Research Service.  There were $5,861$ labels, ranging
from \emph{World Wide Web} to \emph{Age}.  We only used issue labels
which were applied to at least twenty five bills in the 12 years under
consideration.  This filter resulted in seventy-four labels which
correspond fairly well to political issues.  These issues, and the
number of documents each label was applied to, is given in
Table~\ref{table:issues}.~

  \begin{figure}[htdp]
  \caption{Issue labels and the number of documents with each label
  (as assigned by the Congressional Research Service) for Congresses 106 to 111 (1999 to
  2010).}
  \small
  \begin{center}
    \begin{tabular}{cc}
    \begin{tabular}{p{6.4cm}p{.7cm}}
    \hline
    \hline
    \textbf{Issue label} & \textbf{No. bills} \\
    \hline
    Women & 25  \\
    Military history & 25  \\
    Civil rights & 25  \\
    Government buildings; facilities; and property & 26  \\
    Terrorism & 26  \\
    Energy & 26  \\
    Crime and law enforcement & 27  \\
    Congressional sessions & 27  \\
    East Asia & 28  \\
    Appropriations & 28  \\
    Business & 29  \\
    Congressional reporting requirements & 30  \\
    Congressional oversight & 30  \\
    Special weeks & 31  \\
    Social services & 31  \\
    Health & 33  \\
    Special days & 33  \\
    California & 33  \\
    Social work; volunteer service; charitable organizations & 33  \\
    State and local government & 34  \\
    Civil liberties & 35  \\
    Government information and archives & 35  \\
    Presidents & 35  \\
    Government employees & 35  \\
    Executive departments & 35  \\
    Racial and ethnic relations & 36  \\
    Sports and recreation & 36  \\
    Labor & 36  \\
    Special months & 39  \\
    Children & 40  \\
    Veterans & 40  \\
    Human rights & 41  \\
    Finance & 41  \\
    Religion & 42  \\
    Politics and government & 43  \\
    Minorities & 44  \\
    Public lands and natural resources & 44  \\
    \hline
  \end{tabular}
   &
    \begin{tabular}{p{6.4cm}p{.7cm}}
    \hline
    \hline
    \textbf{Issue label} & \textbf{No. bills} \\
    \hline
    Europe & 44  \\
    Military personnel and dependents & 44  \\
    Taxation & 47  \\
    Government operations and politics & 47  \\
    Postal facilities & 47  \\
    Medicine & 48  \\
    Transportation & 48  \\
    Emergency management & 48  \\
    Sports & 52  \\
    Families & 53  \\
    Medical care & 54  \\
    Athletes & 56  \\
    Land transfers & 56  \\
    Armed forces and national security & 56  \\
    Natural resources & 58  \\
    Law & 60  \\
    History & 61  \\
    Names & 62  \\
    Criminal justice & 62  \\
    Communications & 65  \\
    Public lands & 68  \\
    Legislative rules and procedure & 69  \\
    Elementary and secondary education & 74  \\
    Anniversaries & 82  \\
    Armed forces & 83  \\
    Defense policy & 92  \\
    Higher education & 103  \\
    Foreign policy & 104  \\
    International affairs & 105  \\
    Budgets & 112  \\
    Education & 122  \\
    House of Representatives & 142  \\
    Commemorative events and holidays & 195  \\
    House rules and procedure & 329  \\
    Commemorations & 400  \\
    Congressional tributes & 541  \\
    Congress & 693  \\
    \hline
  \end{tabular}
  \end{tabular}
  \end{center}
  \label{table:issues}
\end{figure}

\subsection{Vocabulary selection}
\label{appendix:vocabulary}
In this section we provide further details of vocabulary selection.

We first converted the words in each bill to a canonical form using
the \emph{Tree-tagger} part-of-speech tagger \citep{treetagger:2008}.
Next we counted all phrases with one to five words.  From these, we
immediately eliminated phrases which occurred in more than $10\%$ of
bills or in fewer than $4$ bills, or which occurred as fewer than
$0.001\%$ of all phrases.  This resulted in a list of 40603 phrases
(called $n$-grams in natural language processing).

We then used a set of features characterizing each word to classify
whether it was good or bad to use in the vocabulary.  Some of these
features were based on corpus statistics, such as the number of bills
in which a word appeared.  Other features used external data sources,
including whether, and how frequently, a word appeared as link text in
a Wikipedia article. We estimated weights for these features using a
logistic regression classifier. To train this classifier, we used a
manually curated list of 458 ``bad'' phrases which were semantically
awkward or meaningless (such as \emph{the follow bill}, \emph{and sec
  amend}, \emph{to a study}, and \emph{pr}).  These were selected as
as negative examples in a $L_2$-penalized logistic regression, while
the remaining words we considered ``good'' words.  We illustrate
weights for these features in \myfig{vocabulary_features}. The best
5,000 phrases under this model were used in the vocabulary.

\begin{figure}
    \begin{tabular}{p{6.2cm}p{7.0cm}p{1.2cm}}
      \hline
      \hline
  \small      \textbf{Coefficient} & \small  \textbf{Summary} & \small  \textbf{Weight} \\
      \hline
      \small
  \small      log(count + 1) & \small  Frequency of phrase in corpus & \small  -0.018 \\
  \small      log(number.docs + 1) & \small  Number of bills containing phrase & \small  0.793 \\
  \small      anchortext.presentTRUE & \small  Occurs as anchortext in Wikipedia & \small  1.730 \\
  \small      anchortext & \small  Frequency of appearing as anchortext in Wikipedia & \small  1.752 \\
  \small      frequency.sum.div.number.docs & \small  Frequency divided by number of bills & \small  -0.007 \\
  \small      doc.sq & \small  Number of bills containing phrase, squared & \small  -0.294 \\
  \small      has.secTRUE & \small  Contains the phrase \emph{sec} & \small  -0.469 \\
  \small      has.parTRUE & \small  Contains the phrase \emph{paragra} & \small  -0.375 \\
  \small      has.strikTRUE & \small  Contains the phrase \emph{strik} & \small  -0.937 \\
  \small      has.amendTRUE & \small  Contains the phrase \emph{amend} & \small  -0.484 \\
  \small      has.insTRUE & \small  Contains the phrase \emph{insert} & \small  -0.727 \\
  \small      has.clauseTRUE & \small  Contains the phrase \emph{clause} & \small  -0.268 \\
  \small      has.provisionTRUE & \small  Contains the phrase \emph{provision} & \small  -0.432 \\
  \small      has.titleTRUE & \small  Contains the phrase \emph{title} & \small  -0.841 \\
  \small      test.pos & \small  $\ln(max(-\mbox{test}, 0) + 1)$ & \small  0.091 \\
  \small      test.zeroTRUE & \small  $1$ if $\mbox{test} = 0$ & \small  -1.623 \\
  \small      test.neg & \small  $\ln(max(\mbox{test}, 0) + 1)$ & \small  0.060 \\
  \small      number.terms1 & \small  Number of terms in phrase is 1 & \small  -1.623 \\
  \small      number.terms2 & \small  Number of terms in phrase is 2 & \small  2.241 \\
  \small      number.terms3 & \small  Number of terms in phrase is 3 & \small  0.315 \\
  \small      number.terms4 & \small  Number of terms in phrase is 4 & \small  -0.478 \\
  \small      number.terms5 & \small  Number of terms in phrase is 5 & \small  -0.454 \\
  \small      log(number.docs + 1) * anchortext & \small  $\ln(\mbox{Number of bills containing phrase})$
  \small      $\times 1_{\{ \mbox{Appears in Wikipedia anchortext} \}}$ & \small  -0.118 \\
  \small      log(count + 1) * log(number.docs + 1) & \small  $\ln(\mbox{Number of bills containing phrase} + 1)$
  \small      $\times \ln(\mbox{Frequency of phrase in corpus} + 1)$ & \small 0.246 \\
      \hline
    \end{tabular}
  \caption{Features and coefficients used for predicting ``good''
  phrases.  Below, $\mbox{test}$ is a test statistic which measures
  deviation from a model assuming that words appear independently;
  large values indicate that they occur more often than expected by
  chance.  We define it as $\mbox{test}=\frac{\mbox{Observed count} -
  \mbox{Expected count}}{\sqrt{\mbox{Expected count under a language
  model assuming independence}}}$.}
  \label{figure:vocabulary_features}
\end{figure}

\end{document}